\documentclass{article}

\usepackage[preprint]{neurips_2026}

\usepackage[utf8]{inputenc} 
\usepackage[T1]{fontenc}    
\usepackage{hyperref}       
\usepackage{url}            
\usepackage{booktabs}       
\usepackage{amsfonts}       
\usepackage{nicefrac}       
\usepackage{microtype}      
\usepackage{xcolor}         

\usepackage{graphicx}
\usepackage{amsmath}
\usepackage{multirow}
\usepackage[table]{xcolor}
\usepackage{caption}
\usepackage[ruled,vlined]{algorithm2e}
\usepackage{colortbl}
\usepackage{placeins}
\usepackage{float}
\usepackage{tcolorbox}
\usepackage[table]{xcolor}
\usepackage{colortbl}
\definecolor{lightgreen1}{RGB}{229,244,238}   
\definecolor{lightyellow1}{RGB}{252,245,224}  
\definecolor{tablegreen}{RGB}{235,242,230}
\definecolor{tableblue}{RGB}{236,242,248}
\definecolor{tablered}{RGB}{255,240,240}
\usepackage{alltt}
\tcbuselibrary{breakable}

\title{R²-Mem: Reflective Experience for Memory Search}

\author{%
  Xinyuan Wang, Wenyu Mao, Junkang Wu, Xiang Wang, Xiangnan He\\
  University of Science and Technology of China\\
  \texttt{wangxinyuan@mail.ustc.edu.cn} \\
}

\begin{document}

\maketitle

\begin{abstract}
Deep search has recently emerged as a promising paradigm for enabling agents to retrieve fine-grained historical information without heavy memory pre-managed. However, existing deep search agents for memory system repeat past error behaviors because they fail to learn from the prior high- and low-quality search trajectories. To address this limitation, we propose \textbf{R²-Mem}, a reflective experience framework for memory search systems. In the offline stage, a \textbf{Rubric-guided Evaluator} scores low- and high-quality steps in historical trajectories, and a \textbf{self-Reflection Learner} distills the corresponding abstract experience. During the online inference, the retrieved experience will guide future search actions to avoid repeated mistakes and maintain high-quality behaviors. Extensive experiments demonstrate that R²-Mem consistently improves both effectiveness and efficiency over strong baselines, improving F1 scores by up to \textbf{22.6\%}, while reducing token consumption by \textbf{12.9\%} and search iterations by \textbf{20.2\%}. These results verify that R²-Mem provides a RL-free and low-cost solution for self-improving LLM agents.
\end{abstract}

\section{Introduction}
The rapid advancement of large language models has allowed AI agents to perform well in many domains. However, this also introduces a key requirement: effective agent memory, which must continuously integrate new information with existing memory~\citep{DBLP:journals/tois/ZhangDBMLCZDW25, hu2026memoryageaiagents, wu2026memoryllmeramodular}. Most existing memory systems rely on pre-managed techniques, such as memory graphs, compressed memory units, or heuristic structures~\citep{zhang2026gmemory, fang2026lightmem, DBLP:conf/ecai/ChhikaraKASY25, DBLP:conf/emnlp/KangJZB25, liu2026simplemem}. In addition, some recent studies further attempt to optimize memory ability through end-to-end reinforcement learning~\citep{yan2026memoryr1enhancinglargelanguage, wang2025memalphalearningmemoryconstruction}. However, both heavily pre-structured memories and end-to-end optimization often lose fine-grained historical details and require high computational cost, limiting their use in long-term retrieval and reasoning~\citep{mao2026bimembidirectionalconstructionhierarchical, hu2026memorymattersmoreeventcentric, yu2026agenticmemorylearningunified}.

To overcome this challenge, recent researchers introduce a new paradigm: \textbf{deep search for memory systems}~\citep{yan2025generalagenticmemorydeep, yuan2025memsearchertrainingllmsreason}, which do not pre-manage or heavily compress historical information in advance. Instead, they perform memory retrieval at runtime through an iterative process of planning, searching and reflection to construct more accurate and task-specific context.

However, iterative memory search introduces a key challenge: existing agents treat each search trajectory in isolation and fail to accumulate reusable experience across trajectories, leading to repeated ineffective planning and reflection behaviors, redundant exploration, degraded retrieval quality, and increased reasoning cost~\citep{yan2025generalagenticmemorydeep, xi2025surveyllmbaseddeepsearch}. This issue is further complicated by the fact that each trajectory contains mixed-quality steps. Even successful trajectories may include unnecessary or incorrect actions, while unsuccessful ones may still contain useful partial behaviors. Therefore, treating a whole trajectory as either good or bad is not reliable for learning from past experiences. Instead, we need a finer-grained way to identify high and low-quality steps for accumulating eusable experience.

\begin{figure}[t]
    \centering
    \includegraphics[width=\linewidth]{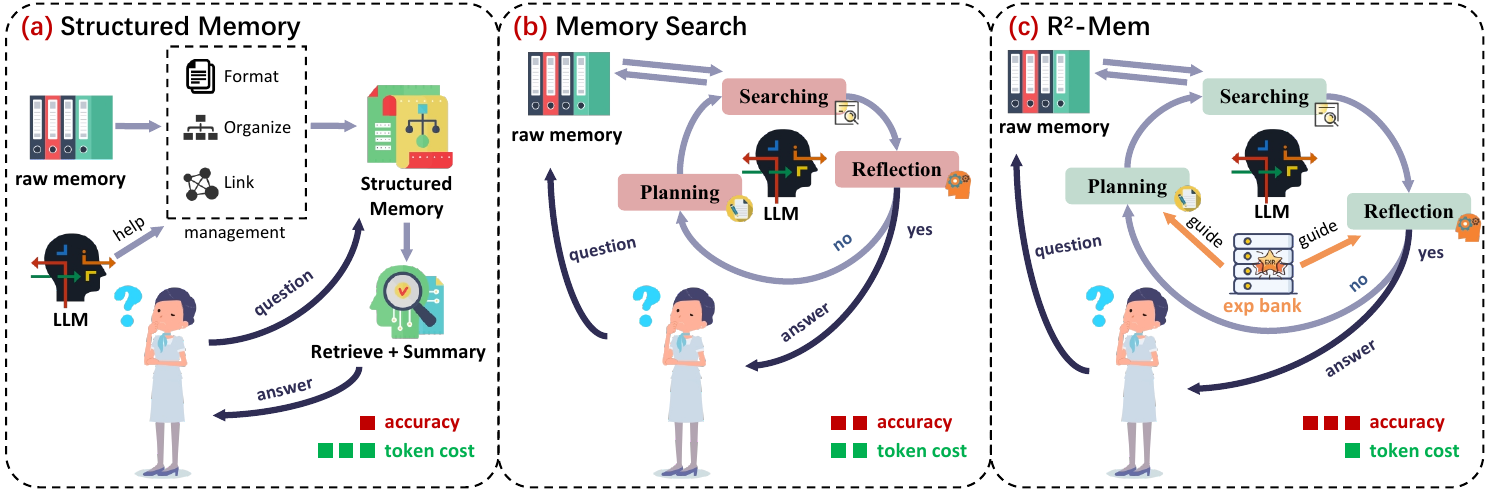}
    \caption{R²-Mem achieves higher scores with lower consumption}
    \label{fig:motivation}
    \vspace{-6mm}
\end{figure}

Motivated by the need for fine-grained step diagnosis and reusable experience accumulation, we propose \textbf{R²-Mem}, a reflective experience framework for memory search. In the offline stage, R²-Mem first employs a \textbf{Rubric-guided Evaluator} to score step-level high- and low-quality behaviors within historical trajectories, and then uses a \textbf{self-Reflection Learner} to distill these behaviors into an experience bank. During online inference, the retrieved experience is reused to guide future planning and reflection, enabling the agent to progressively avoid repeated mistakes while maintaining high-quality search behaviors. Compared to traditional memory architectures and memory search systems(illustrate in Figure~\ref{fig:motivation}), R²-Mem achieves higher scores with lower consumption.

Experiments on three long-context memory benchmarks show that R²-Mem consistently outperforms strong baselines in both retrieval quality and computational efficiency, demonstrating a RL-free and low-cost enhancement for memory search systems.

\section{Related Work}
\subsection{Iterative Deep Memory Search for LLM Agents}
Recent studies have shifted agent memory from pre-managed memory systems\citep{shu2026tracememweavingnarrativememory, xu2026amem, fang2026lightmem, zhang2026gmemory}, toward iterative deep search memory systems, where agents dynamically conduct multi-step planning, searching, and reflection over raw historical contexts to progressively identify question-relevant evidence~\citep{yan2025generalagenticmemorydeep, yuan2025memsearchertrainingllmsreason}. By avoiding aggressive memory compression and heuristic pre-structuring, this paradigm preserves richer fine-grained historical details and demonstrates strong effectiveness for long-context retrieval and reasoning.

Nevertheless, iterative memory search also produces long search trajectories containing abundant high-quality and low-quality planning/reflection behaviors. Existing systems mainly use these trajectories only for the current retrieval episode, without mining them into reusable guidance for subsequent searches. Consequently, similar ineffective exploration behaviors can repeatedly emerge across different queries, resulting in redundant search iterations and degraded accuracy. Our work addresses this underexplored limitation by enabling deep search agent for memory systems to accumulate reflective search experience from prior trajectories.

\subsection{Experience Learning and Agent Self-Evolution}
Another closely related research line studies how LLM agents can improve through accumulated experience from prior interactions~\citep{ouyang2026reasoningbank, ye2026onlineexperientiallearninglanguage, zhang2026expseekselftriggeredexperienceseeking, zhang2025agentlearningearlyexperience}. These approaches typically externalize successful trajectories, verbal reflections, or reusable skills into an experience repository for future retrieval, demonstrating that agents can partially self-evolve without expensive end-to-end policy retraining~\citep{zhang2026memskilllearningevolvingmemory, chen2026scaling}.

However, most existing experience-learning methods operate at the trajectory or task level, treating each historical interaction as a coarse reusable unit. Such granularity is insufficient for iterative deep memory search, where final retrieval quality is often governed by planning and reflection actions within long multi-step search trajectories. In contrast, R²-Mem focuses on distilling \emph{step-conditional reflective experience} from fine-grained high- and low-quality intermediate behaviors, enabling more targeted corrective guidance in future iterative retrieval.

\subsection{Rubric-based Fine-grained Critique Signals}

Rubric-based supervision has recently been explored as a structured alternative to scalar rewards by decomposing model outputs into explicit evaluation dimensions~\citep{lv2026learningqueryspecificrubricshuman, wei2025evomemorybenchmarkingllmagent, shen2026rethinkingrubricgenerationimproving}. Existing studies mainly employ rubric-guided signals for final-result assessment, reward construction, or RL alignment, where the primary goal is to provide more controllable supervision over holistic answer quality~\citep{huang2025reinforcementlearningrubricanchors, jia2025autorubricr1vrubricbasedgenerativerewards, liu2026openrubricsscalablesyntheticrubric}.

In contrast, rubric signals have been far less studied for diagnosing intermediate process quality within long multi-step trajectories, and remain largely unexplored in iterative memory search scenarios. However, the performance of deep search agent in memory systems is often highly sensitive to planning and reflection actions. Motivated by this observation, we introduce rubric-guided critiques as a fine-grained evaluator to identify step-level high- and low-quality search behaviors, which further enables stable reflective experience abstraction from historical trajectories.

\section{Preliminary}
\subsection{Deep Memory Search System}
Recent advances in agent systems have introduced several novel memory architectures. deep memory search systems performs better. Deep  Memory Search System is conceptualized as a process of deep search. It can be regarded as a cycle of \textbf{Planning}, \textbf{Searching} and \textbf{Reflection}.

\begin{figure}[h]
    \vspace{-3mm}
    \centering
    \includegraphics[width=0.66\linewidth]{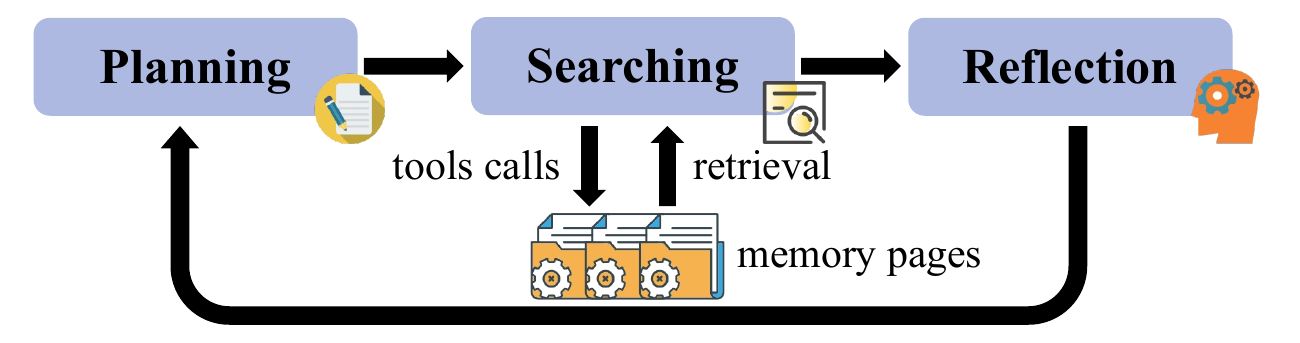}
    \caption{Deep Memory Search System}
    \label{fig:memory_search}
    \vspace{-3mm}
\end{figure}

The overall framework of the memory search system is shown in Figure~\ref{fig:memory_search}.
Formally, the memory search process is formulated as an iterative procedure initialized with the original question $Q$. Specifically, the initial current query is set as $q_1$ = $Q$. for iteration $i$ is defined as follows: Let $f_{\text{plan}}$, $f_{\text{search}}$, $f_{\text{integrate}}$, and $f_{\text{reflect}}$ denote the planning, searching, integration, reflection, respectively.\\
\textbf{Planning}: Given a current question $q_i$ and current memory $m_i$, the system first performs structured reasoning to identify explicit information needs. It decomposes the question $q_i$ into concrete sub-questions-needs $\{q_{i1}, q_{i2}, \dots, q_{in}\}$ and generates a tool-aware retrieval plan $p_i$, specifying which retrieval strategies (e.g., keyword, semantic, or page-level) should be used.
Overall, this process of Planning can be described as:
\begin{equation}
f_{\text{plan}}(q_i, m_i) \rightarrow \quad p_i \in \{\text{keyword}, \text{semantic}, \text{page}\}
\end{equation}

\textbf{Searching}: Based on the generated plan $p_i$, the system executes retrieval actions to collect relevant memory from the memory store $M$ by tools. Retrieved results $\{r_{i1}, r_{i2}, \dots, r_{in}\}$ are aggregated and integrated into a unified intermediate temp memory $m_{i+1}$ that accumulates relevant information:
\begin{equation}
f_{\text{search}}(p_i, M) \rightarrow \{r_{i1}, r_{i2}, \dots, r_{in}\}
\end{equation}
\begin{equation}
f_{\text{integrate}}(q_i, m_i, \{r_{i1}, r_{i2}, \dots, r_{in}\}) \rightarrow m_{i+1}
\end{equation}

\textbf{Reflection}: After each round of planning and searching, the system evaluates whether the collected temp memory $m_{i+1}$ sufficiently satisfies the original question $Q$. If critical information is missing, it formulates refined follow-up queries $q_{i+1}$ and triggers another research iteration. The process continues until information completeness is achieved.
\begin{equation}
f_{\text{reflect}}(m_{i+1}, Q) = \text{enough}, \quad
\text{output} =
\begin{cases}
m_{i+1}, & \text{if enough == true} \\
\text{DeepSearch}(q_{i+1}, m_{i+1}), & \text{otherwise}
\end{cases}
\end{equation}
This closed-loop design enables adaptive information acquisition and allows computation to scale with task complexity.

\subsection{Memory Search Trajectory}
Given an initial user query $Q$, the deep memory search process proceeds through multiple iterative interactions with the memory store, To capture this evolving search behavior, we define a \textit{memory search trajectory} as the ordered sequence of intermediate search steps:
\begin{equation}
\tau = (s_0, s_1, \dots, s_T)
\end{equation}
where each step $s_i$ is represented as
\begin{equation}
s_i = (q_i, \text{action}_i, m_i),
\end{equation}
with $q_i$ denoting the current query at iteration $i$, $\text{action}_i$ denoting the corresponding search actions(planning, searching, reflection), and $m_i$ denoting the temporary memory state updated from retrieved results.

The effectiveness of deep search depends on the quality of intermediate actions in the search trajectory. However, existing systems usually evaluate retrieval performance only based on final results, while errors in intermediate planning and reflection steps are often hard to observe. Since incorrect decisions made in early steps may affect later iterations, they can lead to accumulated errors and inefficient search. Therefore, improving deep memory search requires both step-level diagnosis of intermediate decisions and the accumulation of useful search experience along the trajectory.

\section{Methodology}
\subsection{Overview of the Framework}
\begin{figure}[h]
    \centering
    \vspace{-2mm}
    \includegraphics[width=\linewidth]{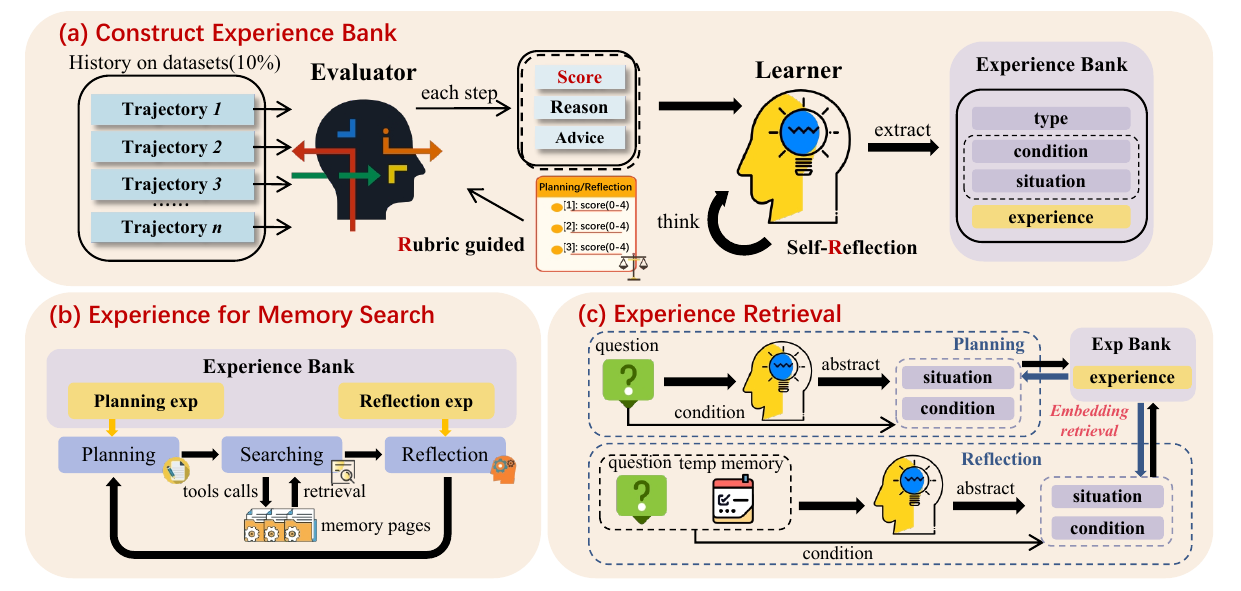}
    \small
    \caption{R$^{2}$-Mem Framework. (a) Construct an experience bank using 10\% of historical trajectories. (b) Perform memory search with experience augmentation on the remaining 90\% of datasets. (c) Detailed process of planning and reflection experience retrieval in our framework.}
    \label{fig:framework}
    \vspace{-3mm}
\end{figure}

We propose \textbf{Rubric-guided Evaluator} and \textbf{self-Reflection Learner} framework:\textbf{R$^{2}$-Mem}, a reflective experience optimization framework for deep Memory Search systems. It introduces a structured process-level diagnosis and accumulation of self-reflection-driven experience to improve memory search precision and search. The overall architecture is illustrated in Figure~\ref{fig:framework}.

The framework consists of two coordinated components: a \textbf{Rubric-guided Evaluator} and a \textbf{self-Reflection-based Learner}. The LLM first performs deep memory search to generate historical search trajectories. The Evaluator then evaluates these trajectories using multi-dimensional rubrics, identifying high- and low-quality behaviors. Based on this diagnosis, the Learner conducts self-reflection to summarize reusable experiences from both high-quality and low-quality trajectories, organizing them into planning-related and reflection-related guidance for future search iterations. Through this Evaluator–Learner interaction, R²-Mem refines search behavior through experiences, reduces redundant exploration, and improves memory retrieval performance.

\subsection{Rubric-guided Evaluating}
Let $\tau = (s_0, s_1, \ldots, s_T)$ denote the trajectory generated during the deep search process for question $Q$, where $s_k$ is the $k$-th step. The rubric list $R=\{d_1, d_2, \dots, d_n\}$ contains multiple evaluation dimensions for assessing the quality of Planning and Reflection. For each step $s_k \in \tau$, the Evaluator first performs dimension-wise assessment over all rubric dimensions and produces a set of fine-grained scores, which are further aggregated into an overall quality score:
\begin{equation}
\text{Evaluator}_{dim}(s_k, R) \rightarrow \{score_{k}^{1}, score_{k}^{2}, \dots, score_{k}^{n}\},
\quad
score_k = \sum_{i=1}^{n} score_{k}^{i}
\end{equation}
where $score_{k}^{i}$ is the score under dimension $d_i$. Based on the aggregated score $score_k$ and the detailed rubric feedback, the Evaluator further generates a corresponding reason $r_k$ and actionable advice $a_k$. Formally, over the entire trajectory:
\begin{equation}
\text{Evaluator}(\tau, R) \rightarrow {(score_k, r_k, a_k)}_{k=0}^{T}
\end{equation}

\subsection{self-Reflection-based experience}
After receiving the corresponding score $score_k$, reason $r_k$, and advice $a_k$ from the Evaluator for a historical trajectory $\tau$, the Learner performs self-reflection to extract useful experience and stores them in the corresponding experience bank. We define the self-reflection process of the Learner as
\begin{equation}
    \text{Learner}_{SR}(s_k, r_k, a_k) \rightarrow e^{t}_{q}, 
    \quad q \in \{\text{good}, \text{bad}\}, 
    \quad t \in \{P, R\}
\end{equation}
where $\text{Learner}_{SR}$ denotes the self-reflection mechanism of the Learner, and $e$ denotes the distilled experience. The quality label $q$ is determined by the evaluation score $score_k$: $q=\text{good}$ if $score_k > K_{\text{high}}$, and $q=\text{bad}$ if $score_k < K_{\text{low}}$. The type label $t$ indicates the functional role of the experience, with $t=P$ for planning-oriented experience and $t=R$ for reflection-oriented experience. Thus, each stored experience is indexed by both quality and function.

\subsection{Deep Search with experience}
\paragraph{Experience bank.}
We maintain two experience banks: a \textit{Planning Experience bank} $\mathcal{D}^P$ and a \textit{Reflection Experience bank} $\mathcal{D}^R$. 
Each bank stores experience pairs of this form:
\begin{equation}
    \mathcal{D} = \{(c_j+\sigma_j, e_j)\}_{j=1}^{N},
\end{equation}
where $c_j$ denotes the condition, $\sigma_j$ represents the abstracted situation, “$+$” denotes the string concatenation. and $e_j$ corresponds to the associated experience.

\paragraph{Condition Construction.}
We denote $Q$ as the overall problem, $q_i$ as the current question, and $m_i$ as the temporary memory at step $i$.
The condition used for experience retrieval depends on the functional role of the experience bank. For planning, the condition is constructed solely from the current query context: $c_i = q_i \quad (Planning).$ For reflection, the condition incorporates both the question $Q$ and the temporary memory: $c_i = [Q + m_i] \quad (Reflection)$.

\paragraph{Situation Construction.}
We assume that $\text{LLM}_\sigma$ is a module used to abstract the current condition $c_i$ into a high-level situational representation, which enables generalization based on the current state.
\begin{equation}
    \text{LLM}_\sigma(c_i) \rightarrow \sigma_i
\end{equation}

\paragraph{Experience Retrieval.}
Given a condition $c_i$, we first abstract it into a high-level situation representation $\sigma_i$ and compute embeddings via $\phi(\cdot)$. We retrieve Top-$K$ most relevant experience from the experience bank $\mathcal{D}$ by similarity search:
\begin{equation}
\mathcal{E}_i^{*} = \operatorname{TopK}
\text{Sim}\big(\phi(c_i + \sigma_i), \phi(c_j + \sigma_j)\big), \quad (c_j, e_j, \sigma_j \in \mathcal{D})
\end{equation}

The retrieved experience are injected into the deep search loop to guide both modules: planning experience $\mathcal{E}_i^{P}$ assist question decomposition and tool-aware action generation $f_{\text{plan}}(q_i, m_i, \mathcal{E}_i^{P})$, while reflection experience $\mathcal{E}_i^{R}$ support sufficiency assessment and next-step request generating $f_{\text{reflect}}(m_{i+1}, Q, \mathcal{E}_i^{R})$.

\section{Experiments}
In this section, we conduct extensive experiments to evaluate the effectiveness, efficiency, and robustness of R$^2$-Mem by answering the following research questions:
\textbf{RQ1:} How does R$^2$-Mem perform on agent memory tasks compared with existing SOTA memory management systems and deep memory search baselines?
\textbf{RQ2:} How does R$^2$-Mem scale with increasing model size, and can it improve search accuracy while reducing search iterations and token cost?
\textbf{RQ3:} To what extent do planning experience and reflection experience individually contribute to the overall effectiveness of R$^2$-Mem?
\textbf{RQ4:} How sensitive is R$^2$-Mem to key hyperparameter settings, including rubrics threshold and experience retrieval number?
\textbf{RQ5:} Can R$^2$-Mem maintain competitive performance when expensive Evaluator supervision is removed and replaced with smaller models? Further details of the experiments are provided in Appendix \ref{sec:app_detailed_experiment}.

\subsection{Experimental Settings}
\textbf{Datasets.} We evaluate R²-Mem on three diverse benchmarks: (1) \textbf{LoCoMo}~\citep{DBLP:conf/acl/MaharanaLTBBF24}, a long-term dialogue dataset requiring multi-hop and temporal reasoning; (2) \textbf{HotpotQA}~\citep{DBLP:conf/emnlp/Yang0ZBCSM18}, a multi-hop QA benchmark based on Wikipedia; and (3) \textbf{NarrativeQA}~\citep{kocisky-etal-2018-narrativeqa}, which involves reasoning over entire books or scripts. 

\paragraph{Baselines.} We compare R$^2$-Mem with four representative families of agent memory systems covering the major existing paradigms. 
(1) \textbf{Memory-free Retrieval}: a vanilla RAG~\citep{DBLP:conf/nips/LewisPPPKGKLYR020} setting that partitions the input into 2,048-token segments and retrieves the top-5 relevant chunks without explicit memory organization. 
(2) \textbf{Structured Memory Management}: A-Mem~\citep{xu2026amem}, MemoryOS~\citep{DBLP:conf/emnlp/KangJZB25}, and LightMem~\citep{fang2026lightmem}, which pre-organize historical information into explicit memory modules for efficient retrieval. 
(3) \textbf{RL-based Memory}: Memory-R1~\citep{yan2026memoryr1enhancinglargelanguage}, which introduces reinforcement learning to optimize memory retrieval behavior.
(4) \textbf{Deep Search Memory}: GAM~\citep{yan2025generalagenticmemorydeep}, which performs iterative planning, searching, and reflection over historical records at runtime. More details of the datasets and baseline implementations are provided in Appendix \ref{sec:app_datasets} and \ref{sec:app_baselines}.

\paragraph{Implementation Details}
We use Qwen2.5 (3B, 7B, 14B)~\citep{bai2023qwentechnicalreport} and llama3.1 (8B) as Learners and GPT-4o as the Evaluator in the overall Performance. BGE-M3~\citep{chen-etal-2024-m3} is used for dense retrieval. For the Locomo datasets, we use 10\% of it to accumulate experience. For HotpotQA and NarrativeQA, we use 20\% of it to accumulate experience. In order to compare with the experimental results of previous scholars and better evaluate the output performance of the model, we use F1 and BLEU1 as our evaluation metrics. We perform inference using Python 3.12 and vLLM on A40 GPUs to obtain our final experimental results.

\subsection{Overall Performance(RQ1)}
Table~\ref{tab:main_results_1} presents the overall results across Qwen2.5-7B and Llama3.1-8B backbones. A clear performance hierarchy can be observed: Memory R1 achieves the best performance on the open-domain category while \textbf{deep-search methods} (GAM) consistently outperform both memory-free and structured-memory approaches, confirming deep memory search as a strong paradigm for long-context reasoning. Notably, \textbf{R$^2$-Mem} further surpasses GAM in almost all settings, with especially significant gains on the challenging LoCoMo benchmark. This consistent improvement over the strongest prior baseline demonstrates the superior effectiveness of R$^2$-Mem in memory-intensive reasoning tasks. In addition, we further conduct supplementary comparisons with GAM on HotpotQA and NarrativeQA, as shown in Table~\ref{tab:main_results_2}, where R$^2$-Mem still maintains competitive advantages. Therefore, we use GAM as the primary baseline for subsequent analysis.

\begin{table}[htp]
\centering
\vspace{-5mm}
\small
\setlength{\tabcolsep}{5pt}
\captionsetup{font=small}
\caption{LoCoMo benchmark results. RL-based Memory R1 shows strong performance on the open-domain category, while the strong baseline GAM achieves competitive results across most task categories. Overall, R$^2$-Mem outperforms all baselines in terms of comprehensive performance.}
\label{tab:main_results_1}
\renewcommand{\arraystretch}{1}
\resizebox{\linewidth}{!}{
\begin{tabular}{llcccccccccc}
\toprule
\textbf{Model} & \textbf{Method}
& \multicolumn{2}{c}{Multi-hop}
& \multicolumn{2}{c}{Temporal}
& \multicolumn{2}{c}{Open domain}
& \multicolumn{2}{c}{Single-hop}
& \multicolumn{2}{c}{\textbf{Overall}} \\
\cmidrule(lr){3-4}
\cmidrule(lr){5-6}
\cmidrule(lr){7-8}
\cmidrule(lr){9-10}
&
& F1 & BLEU
& F1 & BLEU
& F1 & BLEU
& F1 & BLEU
& \textbf{F1} & \textbf{BLEU} \\
\midrule

\multirow{8}{*}{\textbf{Qwen2.5-7B}}
& RAG       & 12.84 & 11.02 
            & 8.65 & 7.94 
            & 8.97 & 6.55 
            & 9.86 & 7.50
            & 10.10 & 8.17 \\
\cmidrule[0.1pt](lr){2-12}
& A-MEM     & 22.59 & 16.28
            & 17.69 & 13.19
            & 14.23 & 12.53
            & 19.96 & 13.56
            & 19.63 & 13.91 \\
& Mem0      & 19.23 & 14.17 
            & 33.26 & 24.68 
            & 17.33 & 11.58 
            & 28.96 & 22.45
            & 27.38 & 20.76 \\
& LIGHTMEM  & 23.45 & 17.61 
            & 32.16 & 26.32 
            & 13.81 & 10.31
            & 32.92 & 26.11
            & 29.90 & 23.67 \\
& MemoryOS  & 26.72 & 17.90 
            & 25.65 & 18.19 
            & 22.05 & 16.95
            & 36.15 & 28.15
            & 31.45 & 23.59 \\

\cmidrule[0.1pt](lr){2-12}
& Memory R1 & 36.55 & 28.71 
            & \cellcolor{lightyellow1}39.05 & 26.03
            & \cellcolor{lightgreen1}\textbf{28.36} & \cellcolor{lightgreen1}\textbf{23.47}
            & 35.64 & 28.06
            & 36.07 & 27.49 \\
\cmidrule[0.1pt](lr){2-12}
& GAM       & \cellcolor{lightyellow1}38.09 & \cellcolor{lightyellow1}29.02
            & 31.10 & \cellcolor{lightyellow1}26.57
            & 23.82 & 20.19
            & \cellcolor{lightyellow1}56.03 & \cellcolor{lightyellow1}49.59
            & \cellcolor{lightyellow1}45.77 & \cellcolor{lightyellow1}39.42 \\
\cmidrule(lr){2-12}
&\textbf{R$^2$-Mem}
            & \cellcolor{lightgreen1}\textbf{41.62} & \cellcolor{lightgreen1}\textbf{32.34}
            & \cellcolor{lightgreen1}\textbf{46.76} & \cellcolor{lightgreen1}\textbf{40.67}
            & \cellcolor{lightyellow1}25.06 & \cellcolor{lightyellow1}20.78 
            & \cellcolor{lightgreen1}\textbf{59.03} & \cellcolor{lightgreen1}\textbf{53.13}
            & \cellcolor{lightgreen1}\textbf{51.35} & \cellcolor{lightgreen1}\textbf{44.91} \\
\midrule

\multirow{8}{*}{\textbf{Llama3.1-8B}}
& RAG       & 13.29 & 10.66 
            & 9.57 & 8.34 
            & 10.58 & 8.31 
            & 12.29 & 9.57
            & 11.81 & 9.44 \\
\cmidrule[0.1pt](lr){2-12}
& A-MEM     & 23.69 & 15.28
            & 18.57 & 13.36
            & 17.59 & 12.23
            & 21.55 & 16.23
            & 21.09 & 15.23 \\
& Mem0      & 20.43 & 15.76 
            & 34.24 & 24.91
            & 16.29 & 10.05 
            & 29.12 & 21.64
            & 27.83 & 20.55 \\
& LIGHTMEM  & 24.42 & 18.66 
            & 31.03 & 25.81
            & 13.31 & 10.61
            & 33.26 & 29.25
            & 30.01 & 25.51 \\
& MemoryOS  & 27.55 & 18.91 
            & 26.22 & 16.96 
            & 23.15 & 15.47
            & 38.43 & 30.15
            & 33.05 & 24.54 \\

\cmidrule[0.1pt](lr){2-12}
& Memory R1 & 35.36 & 26.43 
            & \cellcolor{lightyellow1}41.26 & 29.04
            & \cellcolor{lightgreen1}\textbf{29.34} & \cellcolor{lightgreen1}\textbf{24.56}
            & 37.02 & 31.23
            & 37.13 & 29.51 \\
\cmidrule[0.1pt](lr){2-12}
& GAM       & \cellcolor{lightyellow1}36.23 & \cellcolor{lightyellow1}27.71
            & 39.66 & \cellcolor{lightyellow1}33.07 
            & 21.04 & 16.37 
            & \cellcolor{lightyellow1}50.88 & \cellcolor{lightyellow1}44.58
            & \cellcolor{lightyellow1}44.15 & \cellcolor{lightyellow1}37.48 \\
\cmidrule(lr){2-12}
&\textbf{R$^2$-Mem}
            & \cellcolor{lightgreen1}\textbf{37.75} & \cellcolor{lightgreen1}\textbf{29.28} 
            & \cellcolor{lightgreen1}\textbf{47.53} & \cellcolor{lightgreen1}\textbf{42.14}
            & \cellcolor{lightyellow1}27.48 & \cellcolor{lightyellow1}22.96 
            & \cellcolor{lightgreen1}\textbf{52.16} & \cellcolor{lightgreen1}\textbf{45.73}
            & \cellcolor{lightgreen1}\textbf{47.13} & \cellcolor{lightgreen1}\textbf{40.66} \\
\bottomrule
\end{tabular}
}
\vspace{-2pt}
\end{table}

\begin{table}[htp]
\centering
\vspace{-2mm}
\small
\setlength{\tabcolsep}{9.5pt}
\renewcommand{\arraystretch}{0.9}
\captionsetup{font=small}
\caption{Comparison with the strongest deep-search baseline GAM on NarrativeQA and HotpotQA. R$^2$-Mem consistently improves F1 across backbones.}
\label{tab:main_results_2}
\resizebox{0.82\linewidth}{!}{
\begin{tabular}{llcccc}
\toprule
\multirow{2}{*}{\textbf{Model}} & \multirow{2}{*}{\textbf{Method}}
& \textbf{NarrativeQA}
& \multicolumn{3}{c}{\textbf{HotpotQA}} \\
\cmidrule(lr){3-3}\cmidrule(lr){4-6}
& & F1 & 56K & 224K & 448K \\
\midrule
\multirow{3}{*}{\textbf{Qwen2.5-3B}}
& GAM        & 20.19 & 36.54 & 31.59 & 28.05 \\
& R$^2$-Mem  & 23.33 & 41.05 & 33.98 & 29.94 \\
& \cellcolor{lightgreen1}\textbf{Improv.}
& \cellcolor{lightgreen1}\textbf{+15.55\%}
& \cellcolor{lightgreen1}\textbf{+12.34\%}
& \cellcolor{lightgreen1}\textbf{+7.57\%}
& \cellcolor{lightgreen1}\textbf{+6.70\%} \\
\midrule
\multirow{3}{*}{\textbf{Qwen2.5-7B}}
& GAM        & 28.56 & 44.47 & 46.63 & 44.61 \\
& R$^2$-Mem  & 31.52 & 51.53 & 47.70 & 44.69 \\
& \cellcolor{lightgreen1}\textbf{Improv.}
& \cellcolor{lightgreen1}\textbf{+10.37\%}
& \cellcolor{lightgreen1}\textbf{+10.89\%}
& \cellcolor{lightgreen1}\textbf{+6.88\%}
& \cellcolor{lightgreen1}\textbf{+5.28\%} \\
\bottomrule
\end{tabular}
}
\vspace{-3mm}
\end{table}

\subsection{Scaling Behavior and Efficiency Analysis (RQ2)}
Table~\ref{tab:main_results_backbones} reports the comparison between R$^2$-Mem and the strongest deep-search baseline GAM across three Qwen2.5 backbones on LoCoMo. We observe that R$^2$-Mem consistently achieves better performance under all model scales, demonstrating that the proposed experience-driven memory mechanism remains effective as the backbone capacity increases. Notably, the improvements are especially evident on smaller and medium-sized models, suggesting that reusable planning and reflection experience can effectively compensate for the limited intrinsic reasoning ability of lightweight LLMs.

\begin{table}[h]
\centering
\vspace{-5mm}
\small
\setlength{\tabcolsep}{2.8pt}
\renewcommand{\arraystretch}{1.1}
\captionsetup{font=small}
\caption{R$^2$-Mem consistently scales better than the strongest deep-search baseline GAM across different Qwen2.5 backbones, while requiring fewer generated tokens and iterations on LoCoMo.}
\label{tab:main_results_backbones}
\resizebox{\linewidth}{!}{
\begin{tabular}{llcccccccccccc}
\toprule
\textbf{Model} & \textbf{Method} 
& \multicolumn{2}{c}{Multi-hop}
& \multicolumn{2}{c}{Temporal}
& \multicolumn{2}{c}{Open Domain}
& \multicolumn{2}{c}{Single-hop}
& \multicolumn{2}{c}{Overall}
& Tokens
& Iterations \\
\cmidrule(lr){3-4} \cmidrule(lr){5-6} \cmidrule(lr){7-8} \cmidrule(lr){9-10} \cmidrule(lr){11-12}
& 
& F1 & BLEU 
& F1 & BLEU
& F1 & BLEU 
& F1 & BLEU
& F1 & BLEU
& 
& \\
\midrule

& GAM 
& 23.90 & 17.80 & 34.78 & 29.62 & 17.53 & 13.67 & 35.16 & 31.20 
& 32.00 & 27.42
& 44.87 & 2.47 \\
& \textbf{R$^2$-Mem}
& \textbf{30.08} & \textbf{23.61} & \textbf{37.44} & \textbf{32.78} & \textbf{20.06} & \textbf{16.65} & \textbf{44.96} & \textbf{39.82}
& \textbf{39.24} & \textbf{34.06}
& 39.07 & 1.97 \\
\rowcolor{blue!6} \cellcolor{white}
\multirow{-3}{*}{\textbf{Qwen2.5-3B}}
& \textit{Gain}
& +25.9\% & +32.6\% & +7.6\% & +10.7\% & +14.4\% & +21.8\% & +27.9\% & +27.6\% 
& +22.6\% & +24.2\%
& -12.9\% & -20.2\% \\
\midrule

& GAM 
& 38.09 & 29.02 & 31.10 & 26.57 & 23.82 & 20.19 & 56.03 & 49.59
& 45.77 & 39.42
& 34.01 & 1.84 \\
& \textbf{R$^2$-Mem}
& \textbf{41.62} & \textbf{32.34} & \textbf{46.76} & \textbf{40.67} & \textbf{25.06} & \textbf{20.78} & \textbf{59.03} & \textbf{53.13}
& \textbf{51.35} & \textbf{44.91}
& 28.93 & 1.52 \\
\rowcolor{blue!6} \cellcolor{white}
\multirow{-3}{*}{\textbf{Qwen2.5-7B}}
& \textit{Gain}
& +9.3\% & +11.4\% & +50.4\% & +53.1\% & +5.2\% & +2.9\% & +5.0\% & +7.1\%
& +12.2\% & +13.9\%
& -14.9\% & -17.4\% \\
\midrule

& GAM 
& 43.89 & 36.31 & 49.39 & 44.14 & 29.25 & 24.05 & 59.83 & 54.14
& 52.99 & 47.09
& 28.85 & 1.55 \\
& \textbf{R$^2$-Mem}
& \textbf{45.45} & \textbf{37.11} & \textbf{53.45} & \textbf{47.85} & \textbf{33.50} & \textbf{29.49} & \textbf{61.59} & \textbf{56.21}
& \textbf{55.36} & \textbf{49.48}
& 27.73 & 1.47 \\
\rowcolor{blue!6} \cellcolor{white}
\multirow{-3}{*}{\textbf{Qwen2.5-14B}}
& \textit{Gain}
& +3.6\% & +2.2\% & +8.2\% & +8.4\% & +14.5\% & +22.6\% & +2.9\% & +3.8\% 
& +4.5\% & +5.1\%
& -3.9\% & -5.2\% \\
\bottomrule
\end{tabular}
}
\vspace{-3mm}
\end{table}

Beyond effectiveness, R$^2$-Mem also shows clear efficiency advantages over GAM. Across all backbones, our method requires fewer generated tokens and fewer reasoning iterations to reach better final answers. This indicates that, instead of repeatedly exploring redundant retrieval paths as in conventional deep search, R$^2$-Mem can leverage accumulated experience to produce more targeted memory planning, thereby reducing both inference cost and search redundancy. A more comprehensive analysis of scalability and efficiency is provided in Appendix \ref{sec:app_Amortized}.

\subsection{Ablation Study (RQ3)}
We analyze the impact of each component in the Evaluator–Learner architecture of R$^2$-Mem. As shown in Table~\ref{tab:ablation_rubric}, removing either the \textit{Evaluator} or the \textit{Learner} consistently degrades performance, confirming that both step-level assessment and experience distillation are essential. In particular, removing the Learner introduces noticeable noise into memory search, as fine-grained signals are no longer properly abstracted into stable reusable experiences. When comparing high- and low-quality experience, we find that learning from low-quality behaviors is more effective, suggesting that failure cases provide stronger corrective signals for improving iterative search. Finally, removing either planning or reflection experience leads to clear performance degradation. Ex indicates that both components are indispensable and play complementary roles in guiding the search process. We present detailed analysis in Appendix \ref{sec:app_Ablation}.

\begin{table}[htp]
\centering
\vspace{-5mm}
\small
\setlength{\tabcolsep}{6.5pt}
\captionsetup{font=small}
\caption{Ablation study on rubric-guided components. ``w/o'' denotes the removal of the module.}
\label{tab:ablation_rubric}
\renewcommand{\arraystretch}{1.1}
\resizebox{\linewidth}{!}{
\begin{tabular}{lcccccccccc}
\toprule
\textbf{Setting} 
& \multicolumn{2}{c}{\textbf{Multi-hop}} 
& \multicolumn{2}{c}{\textbf{Temporal}} 
& \multicolumn{2}{c}{\textbf{Open Domain}} 
& \multicolumn{2}{c}{\textbf{Single-hop}} 
& \textbf{Tokens} & \textbf{Iterations} \\
\cmidrule(lr){2-3} \cmidrule(lr){4-5} \cmidrule(lr){6-7} \cmidrule(lr){8-9}
& F1 & BLEU & F1 & BLEU & F1 & BLEU & F1 & BLEU & & \\
\midrule

w/o Evaluator
& 37.47 & 28.31 & 35.12 & 27.84 & 23.88 & 20.25 & 55.74 & 48.36 & 32.41 & 1.65 \\

w/o Learner
& 36.26 & 29.02 & 32.69 & 25.51 & 22.14 & 19.19 & 56.36 & 49.28 & 35.01 & 1.88 \\

\midrule

only high-quality
& 37.36 & 28.84 & 45.59 & 39.50 & 23.15 & 19.77 & 57.29 & 51.49 & 29.11 & 1.55 \\

only low-quality
& 39.33 & 30.27 & \underline{46.63} & \underline{40.39} & 24.35 & 20.21 & \underline{59.00} & \underline{53.04} & \underline{28.98} & 1.57 \\

\midrule

w/o Planning 
& 37.96 
& 29.61 
& 45.61 
& 39.04 
& 24.78 
& \underline{20.63} 
& 58.92
& 52.87
& 28.99 
& \underline{1.54} \\

w/o Reflection   
& \underline{40.00} 
& \underline{31.18} 
& 46.02
& 39.44
& \textbf{26.03} 
& 20.58 
& 58.07 
& 52.24 
& 29.44 
& 1.56 \\
\midrule

\rowcolor{tableblue} \textbf{R$^2$-Mem (Full)} 
& \textbf{41.62} 
& \textbf{32.34} 
& \textbf{46.76} 
& \textbf{40.67} 
& \underline{25.06} 
& \textbf{20.78} 
& \textbf{59.03} 
& \textbf{53.13} 
& \textbf{28.93} 
& \textbf{1.52} \\
\bottomrule
\end{tabular}
}
\vspace{-5mm}
\end{table}

\subsection{Hyperparameter Analysis(RQ4)}
\begin{figure}[h]
    \centering
    \vspace{-4mm}
    \includegraphics[width=\linewidth]{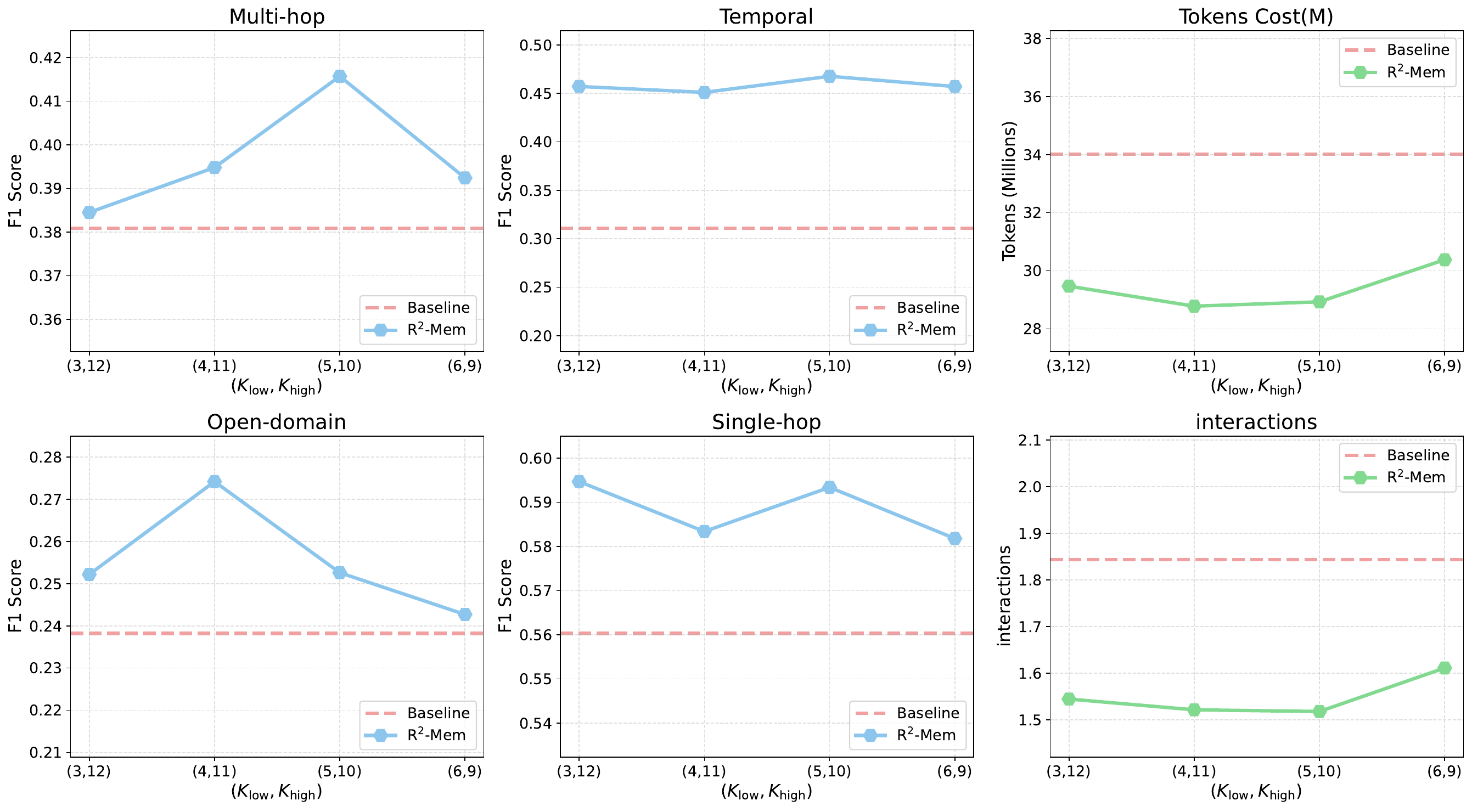}
    \caption{R$^2$-Mem remains consistently better than the baseline across all settings.}
    \label{fig:rubrics}
    \vspace{-4mm}
\end{figure}

\paragraph{Rubrics Threshold Selection for Trajectory Quality.}
We analyze the effect of trajectory filtering thresholds by varying $(K_{\text{low}}, K_{\text{high}})$ within a reasonable range, as shown in Figure~\ref{fig:rubrics}. Here, $K_{\text{low}}$ and $K_{\text{high}}$ define the selection criteria over rubric scores: trajectories with scores below $K_{\text{low}}$ are treated as low-quality steps, while those above $K_{\text{high}}$ are considered high-quality steps.

Overall, R$^2$-Mem is robust to moderate variations of these thresholds, showing stable performance across most settings. In contrast, performance degrades at extreme configurations such as $(6,9)$ and $(3,12)$, where filtering becomes either overly permissive or strict, leading to noisy or insufficient training signals. This indicates that the method is insensitive to threshold within a reasonable range.

\begin{figure}[h]
    \centering
    \includegraphics[width=\linewidth]{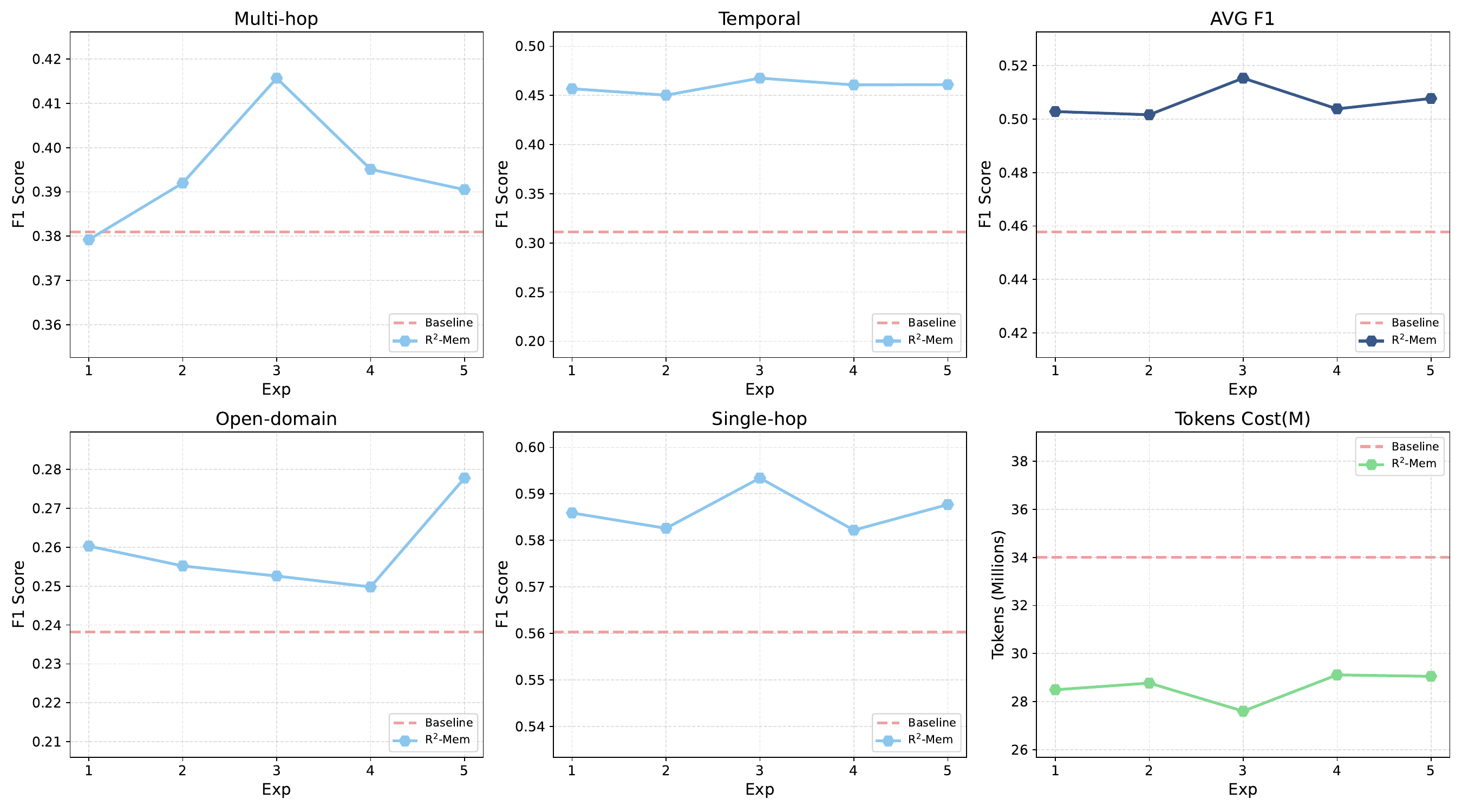}
    \caption{Effect of experience retrieval size $k$ on model performance and retrieval quality.}
    \label{fig:experience}
    \vspace{-5mm}
\end{figure}

\paragraph{Experience Retrieval.}
We further study the impact of the retrieval size $k$ for experience selection, as shown in Figure~\ref{fig:experience}. Results show that performance is stable under moderate values of $k$. Smaller values provide limited guidance, while larger values may introduce redundancy and noise. This indicates a natural trade-off between information richness and retrieval quality. Overall, R$^2$-Mem is not highly sensitive to $k$, and a small number of retrieved experiences is sufficient to achieve strong performance. Detailed studies are provided in Appendix \ref{sec:app_Parameter}.

\subsection{Self-Evolution Capability (RQ5)}
We study the self-evolution capability of R$^2$-Mem by replacing the Evaluator with the Learner itself, forming a self-learning loop. This setting evaluates whether the model can reliably critique and improve its own trajectories without stronger external supervision.
In the standard setting, trajectories generated by the Learner (Qwen2.5-7B) are evaluated by a stronger model (GPT-4o). We instead construct a fully self-evolving variant by using Qwen2.5-7B as both Learner and Evaluator, forming a self-evolving style loop. We report further analysis of self-evolution capability in Appendix \ref{sec:app_selfevo}.

\begin{figure}[h]
\vspace{-3mm}
\centering
\includegraphics[width=\linewidth]{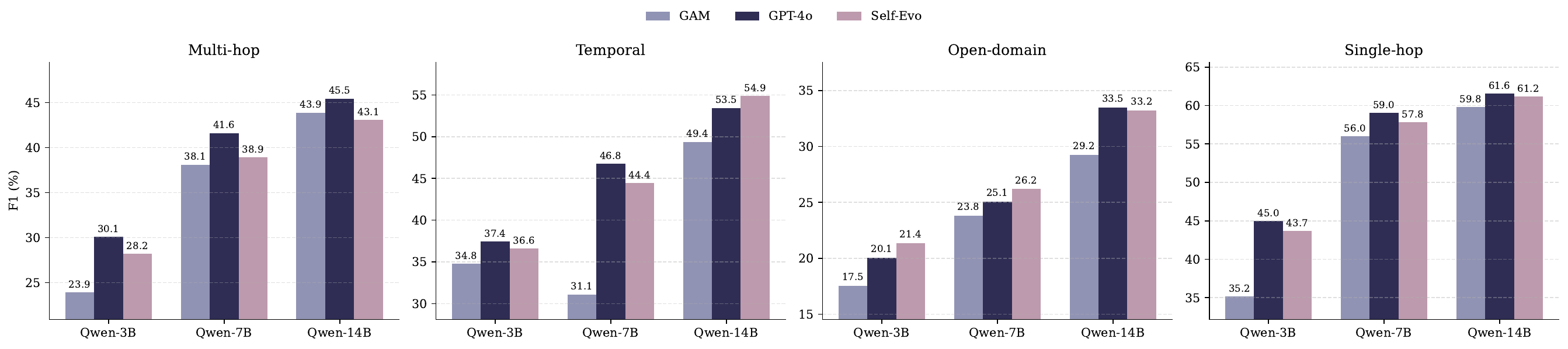}
\caption{Self-evolution performance of R$^2$-Mem under different backbone models.}
\label{fig:self-evo}
\vspace{-3mm}
\end{figure}

In Figure~\ref{fig:self-evo}, Self-Evo outperforms GAM across backbones and tasks. Compared with GPT-4o-based evaluation, it achieves competitive results, reaching near parity on smaller models and maintaining a small gap on larger ones, demonstrating strong self-evaluating without strong external guidance.

\section{Conclusion}
We identified key limitations of deep memory search systems, including their reliance on outcome-level supervision and the lack of experience accumulation. To address these issues, we proposed R$^2$-Mem, a process-aware framework that combines Rubric-guided step-level evaluation with Reflection-based experience reuse. Experiments show that R$^2$-Mem improves memory search performance while reducing search iterations and token consumption. Notably, it also exhibits self-evolution capability by enabling the model to refine its own trajectories without relying on stronger external supervision. These results highlight the importance of process-level supervision for building more efficient and reliable memory-augmented agents.

\begin{ack}
Use unnumbered first level headings for the acknowledgments. All acknowledgments
go at the end of the paper before the list of references. Moreover, you are required to declare
funding (financial activities supporting the submitted work) and competing interests (related financial activities outside the submitted work).
More information about this disclosure can be found at: \url{https://neurips.cc/Conferences/2026/PaperInformation/FundingDisclosure}.

Do {\bf not} include this section in the anonymized submission, only in the final paper. You can use the \texttt{ack} environment provided in the style file to automatically hide this section in the anonymized submission.
\end{ack}

\bibliographystyle{plainnat}
\bibliography{References}


\appendix
\newpage
\section{Algorithm}

\begin{algorithm}[H]
\caption{Construct Experience Banks}
\label{alg:experience_bank}
\KwIn{Historical trajectory $\tau = (s_0, s_1, \dots, s_T)$, rubric list $R$, thresholds $K_{\text{high}}, K_{\text{low}}$}
\KwOut{Planning and reflection experience banks $\mathcal{D}^P, \mathcal{D}^R$}

\For{$k = 0$ \KwTo $T$}{
    $(score_k, r_k, a_k) \leftarrow \text{Evaluator}(s_k, R)$ \;
    \tcp*{rubric-guided evaluation}

    \If{$score_k > K_{\text{high}}$}{
        $q \leftarrow \text{good}$ \;
    }
    \ElseIf{$score_k < K_{\text{low}}$}{
        $q \leftarrow \text{bad}$ \;
    }
    \Else{
        continue \;
    }

    \ForEach{$t \in \{P, R\}$}{
        $e_q^t \leftarrow \text{Learner}_{SR}(s_k, r_k, a_k, t)$ \;
        \tcp*{experience accumulation}

        $c_k^t \leftarrow \text{Cond}(s_k, t),\ \sigma_k^t \leftarrow \text{LLM}_{\sigma}(c_k^t)$ \;
        $\mathcal{D}^t \leftarrow \mathcal{D}^t \cup \{(c_k^t+\sigma_k^t, e_q^t)\}$ \;
        \tcp*{indexed experience storage}
    }
}
\Return $\mathcal{D}^P, \mathcal{D}^R$ \;
\end{algorithm}

\begin{algorithm}[H]
\caption{Deep Search with Experience Augmentation}
\label{alg:deepsearch_exp}
\KwIn{Question $Q$, memory store $M$, planning bank $\mathcal{D}^P$, reflection bank $\mathcal{D}^R$}
\KwOut{Final answer $ans$}

Initialize $q_1 \leftarrow Q$, $m_1 \leftarrow \emptyset$, $i \leftarrow 1$ \;

\While{True}{

    $c_i^P \leftarrow q_i,\ \sigma_i^P \leftarrow \text{LLM}_{\sigma}(c_i^P)$ \;
    $\mathcal{E}_i^P \leftarrow \operatorname{TopK}
    \text{Sim}\big(\phi(c_i^P+\sigma_i^P), \phi(c_j+\sigma_j)\big),
    \ (c_j+\sigma_j,e_j)\in\mathcal{D}^P$ \;
    \tcp*{planning experience retrieval}

    $(\{q_{i1}, \dots, q_{in}\}, p_i) \leftarrow f_{\text{plan}}(q_i, m_i, \mathcal{E}_i^P)$ \;
    $R_i \leftarrow f_{\text{search}}(p_i, M)$ \;
    $m_{i+1} \leftarrow f_{\text{integrate}}(q_i, m_i, R_i)$ \;
    \tcp*{experience-guided planning and memory update}

    $c_i^R \leftarrow [Q + m_{i+1}],\ \sigma_i^R \leftarrow \text{LLM}_{\sigma}(c_i^R)$ \;
    $\mathcal{E}_i^R \leftarrow \operatorname{TopK}
    \text{Sim}\big(\phi(c_i^R+\sigma_i^R), \phi(c_j+\sigma_j)\big),
    \ (c_j+\sigma_j,e_j)\in\mathcal{D}^R$ \;
    \tcp*{reflection experience retrieval}

    $flag \leftarrow f_{\text{reflect}}^{stop}(m_{i+1}, Q, \mathcal{E}_i^R)$ \;
    \If{$flag = \text{true}$}{
        break \;
    }

    $q_{i+1} \leftarrow f_{\text{reflect}}^{next}(q_i, m_{i+1}, \mathcal{E}_i^R)$ \;
    \tcp*{experience-guided reflection}

    $i \leftarrow i+1$ \;
}
$ans \leftarrow f_{\text{answer}}(Q, m_{i+1})$ \;
\Return $ans$ \;
\end{algorithm}

\section{Detailed Experimental Settings}
\label{sec:app_detailed_experiment}
Our detailed experimental results and code are available at: \url{https://github.com/NeurIPS-2026-code/Reflective-Experience-for-Memory-Search}.

\subsection{Details of Datasets}
\label{sec:app_datasets}
\paragraph{LoCoMo.} LoCoMo is a comprehensive benchmark designed to evaluate an agent's capability to maintain, retrieve, and reason over information throughout extended, multi-session dialogues. Moving beyond traditional short-context QA, LoCoMo emphasizes long-term memory persistence in realistic conversational settings. It necessitates that models perform not only direct fact recall but also complex compositional retrieval and temporal reasoning. Following prior work, our experiments adopt four representative task types provided by LoCoMo, including , \textit{multi-hop retrieval(cate.1)}, \textit{temporal reasoning(cate.2)}, \textit{open-domain question answering(cate.3)}, and \textit{single-hop retrieval(cate.4)}. To evaluate the quality of memory retrieval, we employ F1 score and BLEU as the primary evaluation metrics.

Each dialogue instance in LoCoMo (denoted as Conv-*) is associated with a collection of evaluation questions spanning different categories. Since the category composition varies across dialogues, the difficulty and evaluation emphasis may differ substantially from one conversation to another. To better understand this variation, we analyze the question category distribution of the ten LoCoMo conversations used in our study, as summarized in Table~\ref{tab:conv_category_distribution}.

\definecolor{lightgray}{gray}{0.9} 
\begin{table}[h]
\centering
\caption{Conversation Questions Distribution}
\label{tab:conv_category_distribution}
\begin{tabular}{c c c c c c}
\toprule
conversation & category 1 & category 2 & category 3 & category 4 & sum \\
\midrule
Conv-26 & 32 & 37 & 13 & 70  & 152 \\
Conv-30 & 11 & 26 & -- & 44  & 81  \\
Conv-41 & 31 & 27 & 8  & 86  & 152 \\
Conv-42 & 37 & 40 & 11 & 111 & 199 \\
Conv-43 & 31 & 26 & 14 & 107 & 178 \\
Conv-44 & 30 & 24 & 7  & 62  & 123 \\
Conv-47 & 20 & 34 & 13 & 83  & 150 \\
Conv-48 & 21 & 42 & 10 & 118 & 191 \\
Conv-49 & 37 & 33 & 13 & 73  & 156 \\
Conv-50 & 32 & 32 & 7  & 87  & 158 \\
\midrule 
\rowcolor{tableblue}
Avg. & 28.2 & 32.1 & 10.7 & 84.1 & 154.0 \\
\bottomrule
\end{tabular}
\end{table}

\begin{figure}[h]
    \centering
    \vspace{-3mm}
    \includegraphics[width=\linewidth]{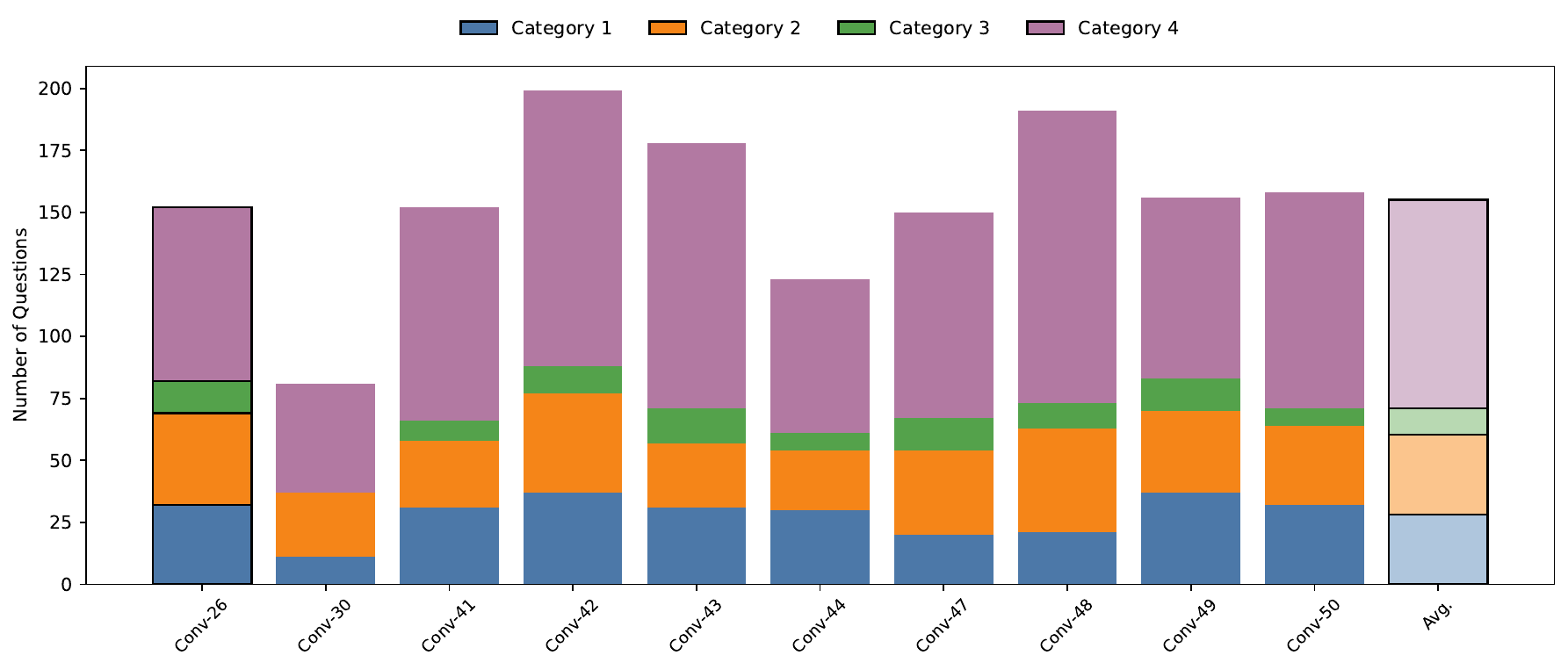}
    \caption{LoCoMo Conversation Category Distribution with Dataset Average}
    \label{fig:locomo_distribution}
    \vspace{-3mm}
\end{figure}

As shown in Table~\ref{tab:conv_category_distribution} and Figure~\ref{fig:locomo_distribution}, the category composition across LoCoMo conversations is not strictly uniform. Some conversations contain a larger proportion of certain reasoning types, while others lack specific categories (e.g., Conv-30 does not contain category-3 questions). This naturally raises the question of whether the trajectory collection source may bias the accumulated experience.

To examine this issue, we conducted preliminary trajectory collection experiments using multiple source conversations, including Conv-26, Conv-41, Conv-47, and Conv-49. Across these settings, we observed comparable downstream performance trends, suggesting that the proposed experience accumulation mechanism is not highly sensitive to the specific choice of source conversation.

For consistency and reproducibility, all main results reported in this paper use Conv-26 as the default trajectory collection conversation. We choose Conv-26 mainly because its category composition and overall question count are close to the dataset average, making it a representative configuration rather than a specially optimized one.

Specifically, we first run standard memory search without R²-Mem on Conv-26 to collect full search trajectories, which are then used to construct step-level planning and reflection experiences. The accumulated experience bank is subsequently reused during inference on the remaining LoCoMo conversations to guide planning and reflection in the deep search process.

\paragraph{HotpotQA.} HotpotQA is a widely used multi-hop question answering benchmark constructed over the Wikipedia corpus, designed to evaluate a model's ability to perform compositional reasoning across multiple supporting documents. In our experiments, we adopt the long-context memory evaluation setting introduced in MemAgent~\citep{yu2026memagent}, where each question is paired with its gold supporting passages along with a varying number of distracting documents. This setting transforms the original benchmark into a memory-intensive retrieval task, requiring the model to identify relevant evidence under substantial contextual interference.

By varying the number of distracting passages, the MemAgent configuration defines three context-length settings, namely 56K, 224K, and 448K tokens, corresponding to the evaluation subsets \texttt{eval\_400}, \texttt{eval\_1600}, and \texttt{eval\_3200}, respectively. To ensure reproducibility, all subsets are generated via random sampling with seed 42.

For each subset, we randomly sample 128 questions as the evaluation pool. Importantly, we further split this pool into two parts to support our process-level optimization framework. The first 20\% (25 questions) is used as a trajectory collection set, where we run standard memory search without R²-Mem to obtain complete search trajectories. The remaining 80\% is reserved for evaluation under the enhanced setting, where the accumulated experience is injected into the memory search process to guide planning and reflection.

This design enables a clear separation between experience construction and performance evaluation, while ensuring that the learned search behavior is grounded in real failure and success cases from the same distribution.

\paragraph{NarrativeQA.} NarrativeQA is a long-context question answering benchmark in which each sample provides an entire book or movie script as supporting context, requiring models to retrieve dispersed evidence and generate answers over extremely long narratives. Compared with document-level QA benchmarks based on shorter passages, NarrativeQA places greater emphasis on sustained context tracking and global memory utilization over tens of thousands of tokens.

In our experiments, we construct the evaluation set through random sampling with seed 42. Specifically, we select 300 questions as the full evaluation pool, with an average context length of approximately 87K tokens. This subset preserves the long-document characteristics of the original benchmark while keeping the evaluation budget manageable for iterative memory search experiments.

Consistent with our HotpotQA setting, we split this pool into two parts to enable experience-driven optimization. 20\% (60 questions) is used for trajectory collection while the remaining 80\% is used for final evaluation.

\subsection{Details of Baselines}
\label{sec:app_baselines}
We provide concise descriptions of all baseline methods for clarity.

\textbf{RAG~\citep{DBLP:conf/nips/LewisPPPKGKLYR020}} performs retrieval-augmented generation by splitting documents into fixed-length chunks (2,048 tokens) and retrieving the top-5 most similar segments via embedding similarity. The retrieved context is directly concatenated with the query for single-pass generation, without explicit memory state or iterative refinement.

\textbf{Mem0~\citep{DBLP:conf/ecai/ChhikaraKASY25}} maintains an external memory store for multi-session conversations by extracting and consolidating salient information from dialogue history. During inference, it retrieves relevant memory entries and uses an LLM-based controller to update or prune stored memories, enabling continuous but heuristic-driven memory management across sessions.

\textbf{A-Mem~\citep{xu2026amem}} organizes interactions into structured memory units inspired by Zettelkasten, where each memory is enriched with attributes such as keywords and contextual descriptions. It constructs a linked memory graph and supports structured retrieval and memory evolution via updates to related entries, but relies on rule-based linking and organization.

\textbf{MemoryOS~\citep{DBLP:conf/emnlp/KangJZB25}} models memory as an operating system with hierarchical levels including short-term, mid-term, and long-term memory. It performs rule-based transitions between levels (e.g., FIFO aggregation and page-based consolidation), enabling structured long-term storage but limiting adaptability due to fixed update policies.

\textbf{GAM~\citep{yan2025generalagenticmemorydeep}} follows a just-in-time memory search paradigm that separates lightweight offline memory from a full page-store. It uses a dual-module design with a Memorizer for compact memory extraction and a Researcher for online retrieval and integration, enabling dynamic context construction during deep search.

\textbf{LightMem~\citep{fang2026lightmem}} introduces a three-stage memory pipeline inspired by human memory, consisting of sensory filtering, topic-aware short-term aggregation, and sleep-time long-term updates. This design reduces redundancy and inference cost through offline consolidation while preserving long-term recall.

\textbf{Memory-R1~\citep{yan2026memoryr1enhancinglargelanguage}} learns memory management policies via reinforcement learning. It adopts a dual-agent design, where a Memory Manager learns structured operations (ADD, UPDATE, DELETE, NOOP) and an Answer Agent retrieves and reasons over memory. Both are optimized using outcome-based RL, enabling adaptive and task-driven memory usage.

\section{More Detailed Experimental Results}

\subsection{Amortized Token Cost of Experience Accumulation}
\label{sec:app_Amortized}
A practical concern for R\textsuperscript{2}-Mem frameworks is whether the process of accumulating experience introduces significant additional token overhead, especially when a stronger evaluator (e.g., GPT-4o) is involved. To address this concern, we systematically analyze token consumption across both the initial experience accumulation stage and subsequent multi-turn conversations, and compare it with the baseline GAM.

\begin{figure}[h]
    \centering
    \vspace{-3mm}
    \includegraphics[width=\linewidth]{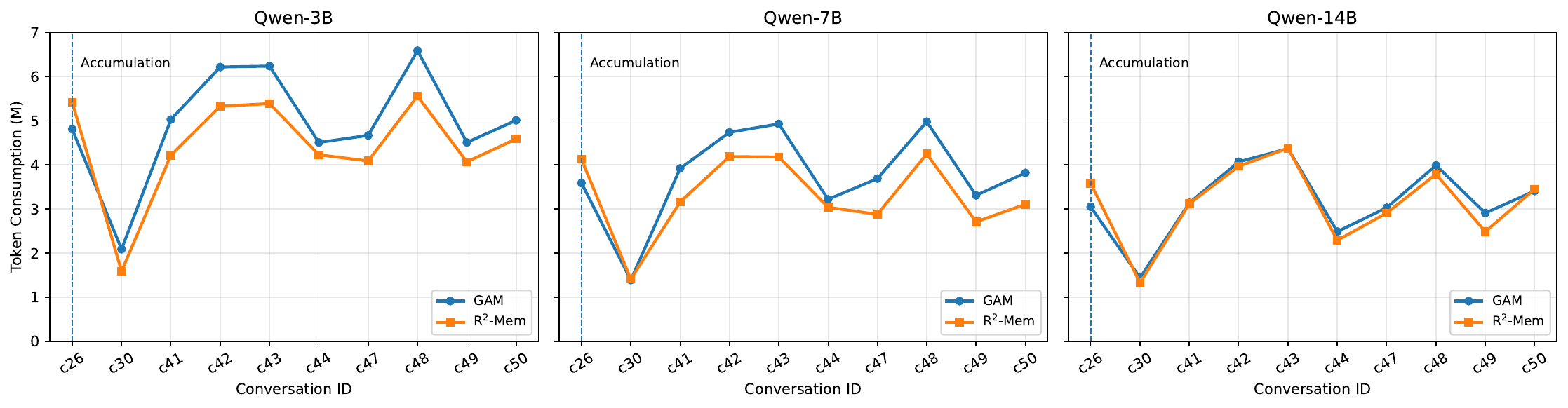}
    \caption{Token consumption across sequential conversations under different backbone models. Conv-26 corresponds to the initial experience accumulation stage.}
    \label{fig:token_amortized}
    \vspace{-3mm}
\end{figure}

We first present the overall token usage trends across sequential conversations under different backbone models in Figure~\ref{fig:token_amortized}. Conv-26 corresponds to the initial experience accumulation stage.

To provide a more fine-grained view of the overall token usage, we further report the full token consumption statistics across all evaluated conversations in Table~\ref{tab:token_full}. The results show that token usage remains stable across different backbones and that R\textsuperscript{2}-Mem consistently achieves lower average token consumption compared to GAM in most settings, indicating that the efficiency gain is not driven by isolated cases but persists throughout downstream tasks.

\begin{table*}[h]
\centering
\small
\setlength{\tabcolsep}{5pt}
\renewcommand{\arraystretch}{1.08}
\captionsetup{font=small}

\caption{Full token consumption statistics (Million) across all evaluated conversations.}
\label{tab:token_full}

\resizebox{\linewidth}{!}{
\begin{tabular}{llccccccccccc}
\toprule
\textbf{Model} & \textbf{Method} & C26 & C30 & C41 & C42 & C43 & C44 & C47 & C48 & C49 & C50 & AVG(9) \\
\midrule

\multirow{2}{*}{Qwen-3B} 
& GAM 
& \cellcolor{lightgreen1}4.81 & 2.09 & 5.03 & 6.22 & 6.24 & 4.51 & 4.67 & 6.59 & 4.51 & 5.01 & 4.986 \\
& R$^2$-Mem 
& \cellcolor{tablered}4.81 + 0.61
& \cellcolor{lightgreen1}1.59 
& \cellcolor{lightgreen1}4.22 
& \cellcolor{lightgreen1}5.33 
& \cellcolor{lightgreen1}5.39 
& \cellcolor{lightgreen1}4.23 
& \cellcolor{lightgreen1}4.09 
& \cellcolor{lightgreen1}5.56 
& \cellcolor{lightgreen1}4.07 
& \cellcolor{lightgreen1}4.59 
& \cellcolor{lightgreen1}4.341 \\

\midrule

\multirow{2}{*}{Qwen-7B} 
& GAM 
& \cellcolor{lightgreen1}3.59 & \cellcolor{lightgreen1}1.39 & 3.92 & 4.74 & 4.93 & 3.22 & 3.69 & 4.98 & 3.31 & 3.82 & 3.778 \\
& R$^2$-Mem 
& \cellcolor{tablered}3.59 + 0.54
& 1.41 
& \cellcolor{lightgreen1}3.16
& \cellcolor{lightgreen1}4.19
& \cellcolor{lightgreen1}4.18
& \cellcolor{lightgreen1}3.04
& \cellcolor{lightgreen1}2.88
& \cellcolor{lightgreen1}4.25
& \cellcolor{lightgreen1}2.71
& \cellcolor{lightgreen1}3.11
& \cellcolor{lightgreen1}3.214\\

\midrule

\multirow{2}{*}{Qwen-14B} 
& GAM 
& \cellcolor{lightgreen1}3.05 & 1.44 & 3.14 & 4.07 & \cellcolor{lightgreen1}4.37 & 2.49 & 3.03 & 3.99 & 2.91 & \cellcolor{lightgreen1}3.41 & 3.206 \\
& R$^2$-Mem 
& \cellcolor{tablered}3.05 + 0.53 
& \cellcolor{lightgreen1}1.33 
& \cellcolor{lightgreen1}3.12 
& \cellcolor{lightgreen1}3.97
& 4.38 
& \cellcolor{lightgreen1}2.29 
& \cellcolor{lightgreen1}2.91 
& \cellcolor{lightgreen1}3.79 
& \cellcolor{lightgreen1}2.49 
& 3.45 
& \cellcolor{lightgreen1}3.081 \\

\bottomrule
\end{tabular}
}

\end{table*}

Having established the overall efficiency trend, we now turn to the initial experience accumulation stage (Conv-26) for a more detailed breakdown. Although this stage involves both an \textbf{Evaluator} and a \textbf{Learner} component, its token cost remains relatively small compared to downstream usage.

\begin{table}[h]
  \centering
  \caption{Amortized token statistics for Conv-26. Conv-26 is decomposed into Evaluator (Eval.) and Learner (Learn.) tokens.}
  \label{tab:token-statistics}
  \begin{tabular}{lccccc} 
    \toprule
    \multirow{2}{*}{\textbf{Model}} & \multicolumn{2}{c}{\textbf{Conv-26}} & \multirow{2}{*}{\textbf{AVG(9) GAM}} & \multirow{2}{*}{\textbf{AVG(9) R\textsuperscript{2}-Mem}} & \multirow{2}{*}{\textbf{Reduction}} \\
    \cmidrule(lr){2-3}
    & \textbf{Eval.} & \textbf{Learn.} & & & \\
    \midrule
    Qwen-3B  & 0.36 & 0.25 & 4.986 & 4.341 & -12.9\% \\
    Qwen-7B  & 0.29 & 0.25 & 3.778 & 3.214 & -14.9\% \\
    Qwen-14B & 0.29 & 0.24 & 3.206 & 3.081 & -3.9\%  \\
    \bottomrule
  \end{tabular}
\end{table}

As shown in Table~\ref{tab:token-statistics}, the Conv-26 stage can be decomposed into an Evaluator (GPT-4o guided) and a Learner (smaller model for experience reflection). Even with the inclusion of a stronger evaluator, the total cost of Conv-26 (approximately 0.29--0.36M tokens) is comparable to only a small number of standard dialogue turns and is incurred only once per task.

More importantly, this one-time experience accumulation is quickly amortized in subsequent usage. The stored experiences are reused across multiple conversations, consistently reducing token costs. As a result, R²-Mem achieves lower average token consumption than GAM in most settings, particularly for Qwen-3B and Qwen-7B, where externalized reasoning guidance leads to more substantial efficiency gains.

Overall, these results demonstrate that the initial experience accumulation introduces only a minor and bounded token overhead, while yielding persistent downstream savings. This confirms that R²-Mem improves efficiency not by shifting cost upward, but by amortizing it effectively over future interactions.

\subsection{Detailed Analysis for Ablation Study}
\label{sec:app_Ablation}
Table~\ref{tab:ablation_rubric} presents a detailed analysis of different ablation settings in R$^2$-Mem, focusing on the roles of the Evaluator–Learner framework and different types of experience construction.

\textbf{(1) Evaluator and Learner.}
We first study two variants: removing the \textit{Evaluator} and removing the \textit{Learner}. 
Removing the Evaluator means that the system no longer performs rubric-based step-level evaluation; instead, the Learner directly summarizes experiences from entire historical trajectories without distinguishing fine-grained step quality. In contrast, removing the Learner means that although step-level signals from trajectories are still identified, they are not distilled into abstract reusable experiences, and are instead directly injected as retrieval context during inference.

Results show that removing either component leads to performance degradation. In particular, without the Learner, the system introduces noticeable noise into memory search, since raw step-level signals cannot be effectively generalized into stable experience representations, weakening long-term reuse.

\textbf{(2) High-quality vs. low-quality experience.}
We further compare two experience construction strategies: using only high-quality steps versus only low-quality steps. The former retains only successful or well-scored behaviors for experience formation, while the latter focuses exclusively on low-scoring or failure cases.

Interestingly, we find that using only low-quality experience performs better, indicating that failure cases provide stronger corrective signals for guiding iterative planning and reflection, and are more informative for improving search robustness.

\textbf{(3) Planning and Reflection experience.}
Finally, we evaluate the effect of removing planning or reflection experience respectively. Removing planning experience eliminates guidance for query decomposition and early-stage search structuring, while removing reflection experience disables feedback signals for evaluating and refining intermediate memory states.

The results show that removing either component leads to clear performance degradation, indicating that planning and reflection experiences are both essential and play complementary roles in guiding effective deep memory search.

Overall, these results validate the importance of fine-grained evaluation, contrastive experience construction, and dual-stage experience guidance in R$^2$-Mem.

\subsection{Detailed Analysis for Parameter Sensitivity}
\label{sec:app_Parameter}
We further provide detailed quantitative results to verify the robustness of R$^2$-Mem under different hyperparameter settings.

\paragraph{Rubric Thresholds.}
Table~\ref{tab:r2mem_rubrics} shows the results under different trajectory filtering thresholds $(K_{\text{low}}, K_{\text{high}})$. 
Overall, the model remains stable across all settings, with Overall F1 varying only from 50.17 to 51.51. 
The best performance is achieved at $(5,10)$, indicating that this setting offers the most balanced trade-off between retaining high-quality trajectories and preserving sufficient diversity.

\begin{table}[h]
\centering
\vspace{-5mm}
\small
\setlength{\tabcolsep}{3pt}
\renewcommand{\arraystretch}{1}
\captionsetup{font=small}
\caption{Parameter sensitivity analysis of R$^2$-Mem with different rubric settings.}
\label{tab:r2mem_rubrics}
\resizebox{\linewidth}{!}{
\begin{tabular}{lcccccccccccc}
\toprule
\textbf{Rubrics} 
& \multicolumn{2}{c}{Multi-hop}
& \multicolumn{2}{c}{Temporal}
& \multicolumn{2}{c}{Open-domain}
& \multicolumn{2}{c}{Single-hop}
& \multicolumn{2}{c}{Overall}
& Interactions
& Tokens(M) \\
\cmidrule(lr){2-3} \cmidrule(lr){4-5} \cmidrule(lr){6-7} \cmidrule(lr){8-9} \cmidrule(lr){10-11}
& F1 & BLEU1
& F1 & BLEU1
& F1 & BLEU1
& F1 & BLEU1
& F1 & BLEU1
& & \\
\midrule

(3,12)
& 38.45 & 29.32
& \cellcolor{lightyellow1}45.71 & 36.38
& 25.22 & \cellcolor{lightyellow1}20.85
& \cellcolor{lightgreen1}59.47 & \cellcolor{lightgreen1}54.24
& \cellcolor{lightyellow1}50.80 & \cellcolor{lightyellow1}44.08
& 1.5448 & 29.47 \\

(4,11)
& \cellcolor{lightyellow1}39.48 & \cellcolor{lightyellow1}30.72
& 45.10 & \cellcolor{lightyellow1}36.95
& \cellcolor{lightgreen1}27.42 & \cellcolor{lightgreen1}23.32
& 58.34 & 52.52
& 50.37 & 43.64
& \cellcolor{lightyellow1}1.5217 & \cellcolor{lightgreen1}28.78 \\

(5,10)
& \cellcolor{lightgreen1}41.57 & \cellcolor{lightgreen1}32.13
& \cellcolor{lightgreen1}46.76 & \cellcolor{lightgreen1}37.91
& \cellcolor{lightyellow1}25.26 & 20.78
& \cellcolor{lightyellow1}59.34 & \cellcolor{lightyellow1}53.51
& \cellcolor{lightgreen1}51.51 & \cellcolor{lightgreen1}44.49
& \cellcolor{lightgreen1}1.5181 & \cellcolor{lightyellow1}28.93 \\

(6,9)
& 39.24 & 30.38
& 45.70 & 35.25
& 24.27 & 19.06
& 58.18 & 52.54
& 50.17 & 42.99
& 1.6114 & 30.38 \\

\bottomrule
\end{tabular}
}
\end{table}

When the threshold is too loose, e.g., $(3,12)$, lower-quality trajectories are more likely to be included, which introduces noise into memory construction. 
In contrast, overly strict filtering such as $(6,9)$ reduces available reasoning diversity, leading to lower performance as well as increased interactions and token consumption. 
These results confirm that R$^2$-Mem is robust to moderate threshold variations and that the default setting is near-optimal.

\paragraph{Exponential Weighting.}
Table~\ref{tab:r2mem_exp} reports the detailed results under different exponential weighting factors. 
We observe similarly stable performance, with Overall F1 fluctuating within a narrow range of 49.35--50.58. 

\begin{table}[h]
\centering
\vspace{-5mm}
\small
\setlength{\tabcolsep}{3pt}
\renewcommand{\arraystretch}{1}
\captionsetup{font=small}
\caption{Parameter sensitivity analysis of R$^2$-Mem under different exponential settings.}
\label{tab:r2mem_exp}
\resizebox{\linewidth}{!}{
\begin{tabular}{lcccccccccccc}
\toprule
\textbf{Exp}
& \multicolumn{2}{c}{Multi-hop}
& \multicolumn{2}{c}{Temporal}
& \multicolumn{2}{c}{Open-domain}
& \multicolumn{2}{c}{Single-hop}
& \multicolumn{2}{c}{Overall}
& Interactions
& Tokens(M) \\
\cmidrule(lr){2-3} \cmidrule(lr){4-5} \cmidrule(lr){6-7} \cmidrule(lr){8-9} \cmidrule(lr){10-11}
& F1 & BLEU1
& F1 & BLEU1
& F1 & BLEU1
& F1 & BLEU1
& F1 & BLEU1
& & \\
\midrule

exp=1
& 37.92 & 28.44
& 45.68 & 36.31
& \cellcolor{lightyellow1}26.03 & \cellcolor{lightyellow1}21.56
& 58.59 & \cellcolor{lightyellow1}52.91
& 49.51 & 43.21
& 1.5357 & \cellcolor{lightyellow1}28.49 \\

exp=2
& 39.20 & 30.25
& 45.03 & 35.21
& 25.52 & 20.95
& 58.26 & 52.38
& 49.55 & 42.98
& 1.5354 & 28.77 \\

exp=3
& \cellcolor{lightgreen1}41.57 & \cellcolor{lightgreen1}32.13
& \cellcolor{lightgreen1}46.76 & \cellcolor{lightyellow1}36.91
& 25.26 & 20.78
& \cellcolor{lightgreen1}59.34 & \cellcolor{lightgreen1}53.51
& \cellcolor{lightgreen1}50.58 & \cellcolor{lightgreen1}44.29
& \cellcolor{lightgreen1}1.5181 & \cellcolor{lightgreen1}27.60 \\

exp=4
& \cellcolor{lightyellow1}39.51 & 29.97
& 46.07 & 35.88
& 24.98 & 19.24
& 58.22 & 52.47
& 49.35 & 43.02
& 1.5419 & 29.11 \\

exp=5
& 39.05 & \cellcolor{lightyellow1}30.58
& \cellcolor{lightyellow1}46.10 & \cellcolor{lightgreen1}37.85
& \cellcolor{lightgreen1}27.78 & \cellcolor{lightgreen1}22.72
& \cellcolor{lightyellow1}58.77 & 52.20
& \cellcolor{lightyellow1}49.69 & \cellcolor{lightyellow1}43.59
& \cellcolor{lightyellow1}1.5224 & 29.05 \\

\bottomrule
\end{tabular}
}
\vspace{-5mm}
\end{table}

Smaller values (exp=1,2) provide insufficient score separation among trajectories, while larger values (exp=4,5) overemphasize a few top-ranked samples and reduce retrieval diversity. 
A moderate setting exp=3 achieves the best overall performance while also requiring the fewest interactions and lowest token usage, suggesting that it provides the most effective balance between trajectory prioritization and memory diversity.

Overall, the results show that R$^2$-Mem does not rely on delicate hyperparameter tuning and maintains stable effectiveness and efficiency across a wide range of settings.

\subsection{Self-Evolution without GPT-4o Evaluation}
\label{sec:app_selfevo}
Table~\ref{tab:locomo_90_final_fixed} provides the detailed quantitative comparison between Self-Evo, GPT-4o-Evaluating R$^2$-Mem, and GAM under different backbone.

\begin{table}[h]
\centering
\vspace{-5mm}
\small
\setlength{\tabcolsep}{5pt}
\captionsetup{font=small}
\caption{Results on 90\% LoCoMo benchmark with GAM as a parallel baseline for each model size}
\label{tab:locomo_90_final_fixed}
\renewcommand{\arraystretch}{1.1}
\resizebox{\linewidth}{!}{
\begin{tabular}{llcccccccccc}
\toprule
\textbf{Model} & \textbf{Method}
& \multicolumn{2}{c}{Multi-hop}
& \multicolumn{2}{c}{Temporal}
& \multicolumn{2}{c}{Open-domain}
& \multicolumn{2}{c}{Single-hop}
& \multicolumn{2}{c}{Overall} \\
\cmidrule(lr){3-4}
\cmidrule(lr){5-6}
\cmidrule(lr){7-8}
\cmidrule(lr){9-10}
&
& F1 & BLEU1
& F1 & BLEU1
& F1 & BLEU1
& F1 & BLEU1
& F1 & BLEU1 \\
\midrule

\multirow{3}{*}{\textbf{Qwen-3B}}
& GAM
& 23.90 & 17.80
& 34.78 & 29.62
& 17.53 & 13.67
& 35.16 & 31.20
& 32.00 & 27.42 \\

& gpt4o
& \cellcolor{lightgreen1}30.08 & \cellcolor{lightgreen1}23.61
& \cellcolor{lightgreen1}37.44 & \cellcolor{lightgreen1}32.78
& \cellcolor{lightyellow1}20.06 & \cellcolor{lightyellow1}16.65
& \cellcolor{lightgreen1}44.96 & \cellcolor{lightyellow1}39.82
& \cellcolor{lightgreen1}39.24 & \cellcolor{lightgreen1}34.06 \\

& \textbf{self-evo}
& \cellcolor{lightyellow1}28.18 & \cellcolor{lightyellow1}21.69
& \cellcolor{lightyellow1}36.61 & \cellcolor{lightyellow1}29.86
& \cellcolor{lightgreen1}21.36 & \cellcolor{lightgreen1}16.98
& \cellcolor{lightyellow1}43.67 & \cellcolor{lightgreen1}40.15
& \cellcolor{lightyellow1}38.09 & \cellcolor{lightyellow1}33.32 \\

\midrule

\multirow{3}{*}{\textbf{Qwen-7B}}
& GAM
& 38.09 & 29.02
& 31.10 & 26.57
& 23.82 & 20.19
& 56.03 & 49.59
& 45.77 & 39.42 \\

& gpt4o
& \cellcolor{lightgreen1}41.62 & \cellcolor{lightgreen1}32.34
& \cellcolor{lightgreen1}46.76 & \cellcolor{lightgreen1}40.67
& \cellcolor{lightyellow1}25.06 & \cellcolor{lightyellow1}20.78
& \cellcolor{lightgreen1}59.03 & \cellcolor{lightgreen1}53.13
& \cellcolor{lightgreen1}51.35 & \cellcolor{lightgreen1}44.91 \\

& \textbf{self-evo}
& \cellcolor{lightyellow1}38.92 & \cellcolor{lightyellow1}30.22
& \cellcolor{lightyellow1}44.44 & \cellcolor{lightyellow1}38.26
& \cellcolor{lightgreen1}26.20 & \cellcolor{lightgreen1}22.96
& \cellcolor{lightyellow1}57.81 & \cellcolor{lightyellow1}51.66
& \cellcolor{lightyellow1}49.76 & \cellcolor{lightyellow1}43.32 \\

\midrule

\multirow{3}{*}{\textbf{Qwen-14B}}
& GAM
& \cellcolor{lightyellow1}43.89 & \cellcolor{lightyellow1}36.31
& 49.39 & 44.14
& 29.25 & 24.05
& 59.83 & 54.14
& 52.99 & 47.09 \\

& gpt4o
& \cellcolor{lightgreen1}45.45 & \cellcolor{lightgreen1}37.11
& \cellcolor{lightyellow1}53.45 & \cellcolor{lightyellow1}47.85
& \cellcolor{lightgreen1}33.50 & \cellcolor{lightgreen1}29.49
& \cellcolor{lightgreen1}61.59 & \cellcolor{lightgreen1}56.21
& \cellcolor{lightgreen1}55.36 & \cellcolor{lightgreen1}49.48 \\

& \textbf{self-evo}
& 43.08 & 34.08
& \cellcolor{lightgreen1}54.88 & \cellcolor{lightgreen1}48.98
& \cellcolor{lightyellow1}33.21 & \cellcolor{lightyellow1}28.67
& \cellcolor{lightyellow1}61.16 & \cellcolor{lightyellow1}56.04
& \cellcolor{lightyellow1}54.93 & \cellcolor{lightyellow1}48.99 \\

\bottomrule
\end{tabular}
}
\vspace{-5mm}
\end{table}

Overall, Self-Evo consistently outperforms GAM under all three backbone sizes, showing that even without GPT-4o as an evaluator, the small model can still evaluate and learn useful experience through self-reflection. In particular, the Overall F1 score improves from 32.00 to 38.09 on Qwen-3B, from 45.77 to 49.76 on Qwen-7B, and from 52.99 to 54.93 on Qwen-14B which demonstrates that our framework does not rely on a stronger evaluate to surpass baselines.

At the same time, Self-Evo remains close to GPT-4o-evaluated R$^2$-Mem. The Overall F1 gap is only 1.15 on Qwen-3B, 1.59 on Qwen-7B, and further shrinks to 0.43 on Qwen-14B. Such a small difference indicates that the backbone model itself is already capable of identifying useful and harmful trajectories, and can convert them into reusable memory experience with limited loss compared with GPT-4o supervision.

In summary, these results show that our method is not strongly dependent on powerful external models. 
Even small backbone models can improve themselves through the proposed memory construction process, achieving clear gains over GAM and performance close to GPT-4o-based evaluation.

\section{Prompt Templates}

In this section, we present the specific prompt templates used in our study, including the Rubric-guided Evaluator and the self-Reflection-based Learner for the offline phase, as well as the experience-based memory search for the online phase.

\subsection{Evaluator}
\begin{tcolorbox}[
    breakable,
    colback=gray!5,
    colframe=blue!50!black,
    left=5pt,
    arc=1pt,
    outer arc=0pt,
    title=Rubric-Guided Evaluator Prompt
]
{\ttfamily
\textbf{System Prompt:} \\
You are an expert evaluator for an AI memory deep search system. \\
Your task is to assess EACH step in the reasoning TRACE. \\
Each step contains two modules: Planning and Reflection. \\
Planning evaluates whether the retrieval strategy is complete, non-redundant, tool-aligned, and efficient. \\
Reflection evaluates whether the system correctly determines if the current evidence is sufficient, whether it avoids unnecessary retrieval, and whether the follow-up request accurately targets missing information. \\

\vspace{0.3em}
\textbf{User Prompt:} \\
QUESTION: \{QUESTION\} \\
REFERENCE\_ANSWER: \{REFERENCE\_ANSWER\} \\
MODEL\_ANSWER: \{MODEL\_ANSWER\} \\
TRACE: \{TRACE\} \\

For each step in TRACE: \\
(1) Evaluate the Planning module using the following rubrics: \\
- Info Needs Coverage (0--3) \\
- Info Needs Non-Redundancy (0--3) \\
- Tool--Info Alignment (0--3) \\
- Planning Efficiency (0--3) \\

(2) Evaluate the Reflection module using the following rubrics: \\
- Sufficiency Judgment Accuracy (0--3) \\
- Minimal Sufficiency Recognition (0--3) \\
- Follow-up Query Quality (0--3) \\
- Answer Completeness Awareness (0--3) \\

For each scored module, provide a concise reason and an abstract advice for future improvement or maintenance. \\
Return ONLY a valid JSON object in the predefined format.
}
\end{tcolorbox}

\begin{tcolorbox}[
    breakable,
    colback=white,
    colframe=black,
    left=5pt,
    arc=1pt,
    outer arc=0pt,
    title=JSON Output Format
]
\begin{alltt}
"results": [\{
        "step": <int number>,
        "module": "Planning" | "Reflection",
        "rubrics": \{
            "Info Needs Coverage": 0-3,
            "Info Needs Non-Redundancy": 0-3,
            "Tool--Info Alignment": 0-3,
            "Planning Efficiency": 0-3
            \},
        "reason and advice": ""\}...
]
\end{alltt}
\end{tcolorbox}

\subsection{self-Reflection Learner}
\paragraph{Planning Experience.} The planning prompt abstracts a detailed step into reusable planning strategies. Given a planning step and its detailed information, the LLM is asked to summarize the planning situation and induce a generalized planning experience.
\begin{tcolorbox}[
    breakable,
    colback=gray!5,
    colframe=blue!50!black,
    left=5pt,
    arc=1pt,
    outer arc=0pt,
    title=Planning Experience Distillation Prompt
]
{\ttfamily
\textbf{System Prompt:} \\
You are an AI TRACE Strategist/Auditor. \\
Your goal is to derive GENERALIZABLE planning experience from judged planning traces. \\
The output experience must follow the form: \\
IF <abstract situation> THEN <retrieval planning strategy>. \\

Do not copy concrete surface facts from the trace. \\
Treat the diagnosed evaluation reason as authoritative supervision. \\
The situation should describe an abstract query type or information gap. \\
The experience should describe how to construct: \\
(1) info\_needs, and (2) retrieval tool selection strategy. \\

\vspace{0.3em}
\textbf{User Prompt:} \\
QUESTION: \{QUESTION\} \\
JUDGED\_PLANNING\_TRACE: \{TRACE\} \\
DIAGNOSED\_REASON: \{REASON\} \\

Your task: \\
1. Analyze why this planning trace is high-quality or low-quality. \\
2. Briefly summarize the trace. \\
3. Abstract the planning situation without using concrete entities. \\
4. Distill a reusable IF--THEN planning experience that can guide future retrieval planning. \\

Return a JSON object with: \\
\{thinking, summary, situation, experience\}
}
\end{tcolorbox}

\begin{tcolorbox}[
    breakable,
    colback=white,
    colframe=black,
    left=5pt,
    arc=1pt,
    outer arc=0pt,
    title=JSON Output Format
]
\begin{alltt}
\{
    "thinking": "<logic analysis grounded in the trace>",
    "summary": "<Briefly describe this Trace(question and plan)>",
    "situation": "<abstract situation>",
    "experience": "IF <abstract situation> THEN <info_need/tool action>"
\}
\end{alltt}
\end{tcolorbox}

\paragraph{Reflection Experience.}
Reflection prompt abstracts a detailed step into reusable decision-making experience which guides sufficiency judgment and generates a new request.

\begin{tcolorbox}[
    breakable,
    colback=gray!5,
    colframe=blue!50!black,
    left=5pt,
    arc=1pt,
    outer arc=0pt,
    title=Reflection Experience Distillation Prompt
]
{\ttfamily
\textbf{System Prompt:} \\
You are an AI Memory TRACE Strategist/Auditor. \\
Your goal is to derive GENERALIZABLE reflection experience from judged reflection traces. \\
The output experience must follow the form: \\
IF <abstract situation> THEN <decision strategy>. \\

Do not copy concrete facts from the trace. \\
Treat the diagnosed evaluation reason as authoritative supervision. \\
The situation should describe an abstract (QUESTION + temp\_memory) state. \\
The experience should describe how to: \\
(1) judge whether current memory is sufficient, \\
(2) decide whether to stop or continue retrieval, and \\
(3) generate a targeted new\_request if needed. \\

\vspace{0.3em}
\textbf{User Prompt:} \\
QUESTION: \{QUESTION\} \\
JUDGED\_REFLECTION\_TRACE: \{TRACE\} \\
DIAGNOSED\_REASON: \{REASON\} \\

Your task: \\
1. Analyze why this reflection trace is high-quality or low-quality. \\
2. Briefly summarize the trace. \\
3. Abstract the reflection situation without using concrete entities. \\
4. Distill a reusable IF--THEN reflection experience that can guide future sufficiency judgment. \\

Return a JSON object with: \\
\{thinking, summary, situation, experience\}
}
\end{tcolorbox}

\begin{tcolorbox}[
    breakable,
    colback=white,
    colframe=black,
    left=5pt,
    arc=1pt,
    outer arc=0pt,
    title=JSON Output Format
]
\begin{alltt}
\{
    "thinking": "<analysis of reflection success/failure>",
    "summary": "<Briefly describe this Trace(question and temp memory)>",
    "situation": "<abstract situation>",
    "experience": "IF <abstract situation> THEN <decision strategy>"
\}
\end{alltt}
\end{tcolorbox}

\section{Trace and Experience Cases}

We first provide a representative memory search trajectory to illustrate the iterative interaction among Planning, temporary memory accumulation, and Reflection-based stopping decisions during online inference.
\subsection{Memory Search Trajectory Example}
\begin{tcolorbox}[
    breakable,
    colback=green!3,
    colframe=green!40!black,
    left=5pt,
    arc=1pt,
    outer arc=0pt,
    title=Memory Search Trajectory Example
]

\textbf{Question:} When did Gina open her online clothing store? \\

\vspace{0.5em}
\textbf{(Step 0)} \\
\textbf{[Planning]}
\begin{itemize}
    \item Info needs: identify exact opening date of Gina's online clothing store
    \item Tools: keyword, vector, page\_index
    \item Keyword collection: 
    "Gina online clothing store launch", 
    "Gina opened online clothing store"
    \item Vector queries: 
    "When did Gina open her online clothing store?"
    \item Page index: 5, 9
\end{itemize}
\textbf{Temp Memory:} 
Gina opened her online clothing store sometime between 29 Jan 2023 and 16 Mar 2023. \\
\textbf{[Reflection] } "enough = False" → generate new request: "What is the exact date when Gina opened her online clothing store? What did Gina do between 29 January, 2023, and 16 March, 2023?" \\

\vspace{0.5em}
\textbf{(Step 1)} \\
\textbf{[Planning]}
\begin{itemize}
    \item Info needs: determine exact opening date
    \item Tools: keyword, vector
    \item Keyword collection: Gina, online clothing store, opened, 29 Jan 2023, 16 Mar 2023
    \item Vector queries: 
    "When did Gina open her online clothing store?", 
    "Find exact launch date of Gina store"
    \item Page index: 5
\end{itemize}
\textbf{Temp Memory:} 
Gina opened her online clothing store on 16 March 2023. \\
\textbf{[Reflection] } "enough = True", new request = NULL → stop retrieval \\

\vspace{0.5em}
\textbf{Final Memory:} Gina opened her online clothing store on 16 March 2023.
\end{tcolorbox}

\subsection{Distilled Experience Cases}
We further provide representative entries from the planning and reflection experience bank.

\begin{tcolorbox}[
    breakable,
    colback=orange!3,
    colframe=orange!60!black,
    left=5pt,
    arc=1pt,
    outer arc=0pt,
    title=Planning Experience Case
]

\textbf{Condition:} How long has Caroline had her current group of friends for? \\

\textbf{Situation:} Time-related query asking for the duration of a relationship or group affiliation. \\

\textbf{Summary:} 
This trace asks about the duration of Caroline's current friendship group, but the original plan decomposes it into multiple separate events instead of directly focusing on the required time span. It applies keyword and vector retrieval to collect evidence about when Caroline met her friends and joined the LGBTQ+ community. \\

\textbf{Distilled Experience:} 
IF the question asks about the duration of a relationship or group affiliation and involves multiple related events, THEN construct a single info\_need centered on the target time frame and use keyword + vector retrieval to locate all relevant dates.

\end{tcolorbox}

\vspace{0.8em}

\begin{tcolorbox}[
    breakable,
    colback=purple!3,
    colframe=purple!60!black,
    left=5pt,
    arc=1pt,
    outer arc=0pt,
    title=Reflection Experience Case
]

\textbf{Condition:} Question: When did Melanie run a charity race? \\
Temp Memory: Melanie ran a charity race for mental health last Saturday. \\

\textbf{Situation:} 
A question asks for a specific factual detail, while the current temp\_memory only provides a coarse temporal reference rather than the exact target information. \\

\textbf{Summary:} 
The question requires the exact date of Melanie's charity race, but the current temp\_memory only states that it happened ``last Saturday,'' which is informative but still not sufficient for answering the question precisely. \\

\textbf{Distilled Experience:} 
IF the question asks for a specific detail and the temp\_memory only contains a general or relative description without that exact detail, THEN set enough = false and generate a new\_request that uses the current coarse clue to retrieve the missing specific information.

\end{tcolorbox}

\section{Discussion and Limitation}
\subsection{Relationship with Skills}
Our experience bank is conceptually related to the notion of \emph{skills} in recent LLM agent studies, since both aim to externalize reusable knowledge beyond the parametric model~\citep{zhang2026memskilllearningevolvingmemory, du2026memoryautonomousllmagentsmechanisms, zhou2026externalizationllmagentsunified, mi2026skillprolearningreusableskills}. 
However, they operate at different levels.

Skill-based methods usually treat a \emph{skill} as a reusable procedural routine, such as a decomposition strategy, a retrieval workflow, or a tool-use policy, which can be invoked as a relatively complete problem-solving module. 
In contrast, the experience stored in R$^{2}$-Mem is not an executable routine, but a lightweight reflective suggestion distilled from a single historical search step under evaluator diagnosis.

Compared with skills, our experiences have three key properties: 
(1) \textbf{step-level granularity}, since they are extracted from detailed Planning or Reflection steps rather than full trajectories; 
(2) \textbf{diagnostic origin}, since they are generated from evaluator-identified success/failure reasons;
(3) \textbf{search-state dependency}, since they provide contextual search adjustments conditioned on the current situation.

Therefore, R$^{2}$-Mem is better viewed as an \emph{experience-guided search refinement framework} rather than a skill execution framework. 
Skills offer coarse-grained reusable task routines, while our experience bank provide fine-grained behavioral guidance during deep memory search.

\subsection{Broader Impact}
This work contributes to the development of more adaptive and reliable long-term memory agents by enabling deep memory search systems to accumulate reusable process-level experience from historical trajectories. 
Through rubric-guided evaluating and self-reflective experience reuse, R$^{2}$-Mem improves search precision and reduces redundant exploration without requiring larger model scales or longer context windows, offering a more efficient path toward robust long-horizon memory search.

The proposed framework may benefit a broad range of memory-intensive applications, including personalized assistants, educational systems, enterprise knowledge agents, and long-context question answering, where stable historical retrieval and process-aware search refinement are essential. 
Moreover, by introducing explicit evaluator diagnosis and interpretable experience guidance, our method provides a more transparent mechanism for optimizing memory search systems, which may facilitate future research on controllable and self-improving LLM memory systems.

\subsection{Limitations}
R$^{2}$-Mem assumes that future search scenarios share some similarity with the historical trajectories used to construct the experience bank. When the downstream distribution differs substantially, retrieved experiences may become less relevant and offer limited guidance.

Moreover, because the experience bank is generated offline, its usefulness is affected by the quality of evaluator judgments and self-reflection. Improving experience robustness under larger distribution shifts is left for future work.

\newpage

\end{document}